\documentclass[journal,twoside,web]{ieeecolor}
\usepackage{generic}
\usepackage{cite}
\let\labelindent\relax
\usepackage{enumitem}
\usepackage{amsmath,amssymb,amsfonts}
\usepackage{algorithmic}
\usepackage{graphicx}
\usepackage{algorithm,algorithmic}
\usepackage[hidelinks]{hyperref}
\usepackage{threeparttable}
\usepackage{textcomp}
\usepackage[font={footnotesize}, labelfont={bf}, textfont={md}]{caption}
\usepackage{lipsum}
\usepackage{gensymb}
\usepackage{xr}
\usepackage{subcaption}
\renewcommand{\baselinestretch}{0.92}

\def\BibTeX{{\rm B\kern-.05em{\sc i\kern-.025em b}\kern-.08em
    T\kern-.1667em\lower.7ex\hbox{E}\kern-.125emX}}
\begin{document}
\title{Characterizing Healthy \& Post-Stroke Neuromotor Behavior During 6D Upper-Limb Isometric Gaming: Implications for Design of End-Effector Rehabilitation Robot Interfaces}
\author{Ajay Anand, \IEEEmembership{Member, IEEE}, Gabriel Parra, Chad A. Berghoff, and Laura A. Hallock,
\IEEEmembership{Member, IEEE}
\thanks{Submitted August 1, 2025. This work was supported by the Department of Mechanical Engineering and the Office of Undergraduate Research at the University of Utah.
The authors are with the Departments of Mechanical Engineering, Physical Medicine \&
Rehabilitation, and Mathematics, the Kahlert School of Computing, and the Robotics Center at the University of Utah, Salt Lake City, UT 84112, USA. Correspondence should be directed to {\tt\footnotesize \{ajay.anand, laura.hallock\}@utah.edu}.
}
\vspace{-2em}}
\maketitle
\begin{abstract}
Successful robot-mediated rehabilitation requires designing games and robot interventions that promote healthy motor practice. However, the 
interplay between a given user’s neuromotor behavior, the gaming interface, and the physical robot makes 
designing system elements --- and even characterizing what 
behaviors 
are 
``healthy'' or pathological --- challenging. 
We leverage our 
OpenRobotRehab~1.0 open access data set 
to assess the characteristics of 13~healthy and 2~post-stroke users' force output, muscle activations, and game performance while executing isometric trajectory tracking tasks using an end-effector rehabilitation robot.
We present 
an
assessment of how subtle aspects of interface design impact user behavior; 
an analysis of how pathological neuromotor behaviors are reflected in 
end effector force dynamics;
and
a novel hidden Markov model (HMM)--based neuromotor behavior classification method based on surface electromyography (sEMG) signals during cyclic motions.
We demonstrate that task specification (including which axes are constrained and how users interpret tracking instructions) 
shapes user behavior; that pathology-related features are detectable in 6D end-effector force data during isometric task execution 
(with significant differences between healthy and post-stroke profiles in force error and average force production at $p=0.05$); 
and that healthy neuromotor strategies are heterogeneous and 
inherently difficult to characterize. We also show that our HMM-based models discriminate healthy and post-stroke neuromotor dynamics where synergy-based decompositions reflect no such differentiation.
Lastly, we discuss these results' implications for the design of adaptive end-effector rehabilitation robots capable of promoting healthier movement strategies across diverse user populations.
\end{abstract}


\begin{figure}[ht]
    \centering
        \includegraphics[width=\linewidth]{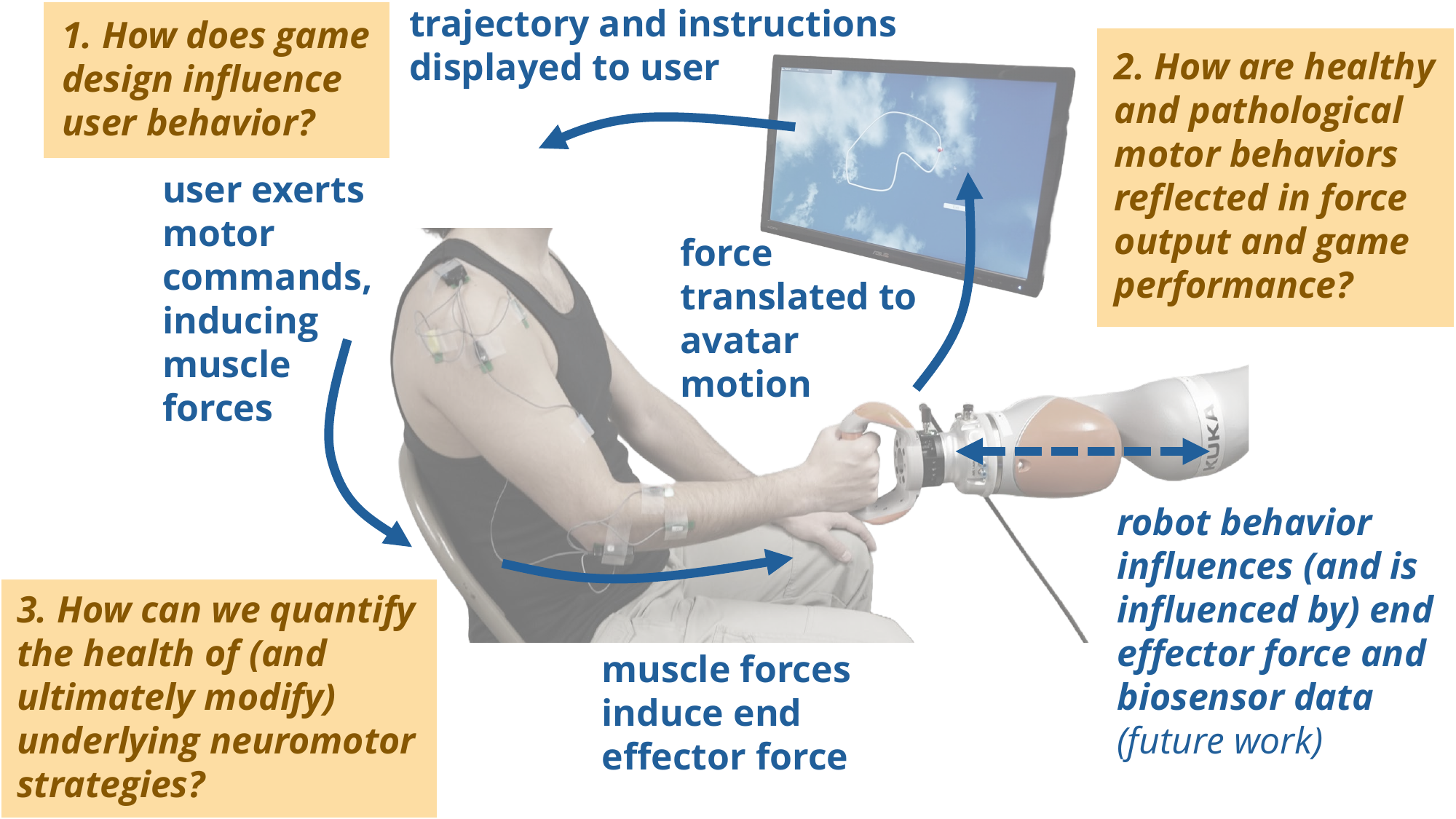}
    \caption{End-effector robot--mediated rehabilitation --- in which users are instructed to perform varied trajectory tracking tasks while assisted (or resisted) by a robot --- consists of complex feedback loops (\emph{blue}) between task specification and display, human neuromotor behavior, and robot intervention. 
    In this work, we leverage multimodal biosensor data, collected from 13 healthy and 2 post-stroke participants during completion of 8 isometric gaming tasks under 2 pose conditions~\cite{anandextensible2025},
    to address three questions key to effective therapy design (\emph{gold}), toward development of rehabilitation robots that can induce targeted, healthy neuromotor practice.%
    }
    \vspace{-2em}
    \label{fig:sum}
\end{figure}
\vspace{-0.5em}
\section{Introduction}
\label{sec:introduction}
Over 2.4 billion individuals worldwide are estimated to live with health conditions that would benefit from neuromuscular rehabilitation \cite{who2024}. Upper-limb motor impairment is particularly prevalent (impacting, for example, an estimated 50--75\% of stroke survivors~\cite{anwer2022rehabilitation}) and especially damaging to the independence and quality of life of those affected. Robot-mediated rehabilitation --- and in particular, end-effector robot rehabilitation, which does not require complicated fitting of exoskeletal elements --- is a promising method to deliver upper-limb therapy~\cite{ellis2009impairment, moulaeiOverviewRoleRobots2023} in a manner that is both personalized and quantifiable 
(and that could help address high costs, long wait times, and clinician scarcity \cite{LIN2015946, wei2024systematic, yagi2017impact}). 
However, despite numerous clinical trials, such robot therapy methods have not yet been shown to induce clinically meaningful differences in impairment measures over usual care \cite{sobreperaAgeMotorFunction2025, park2024effects}, and patients often retain substantial neuromotor deficits regardless of therapy approach \cite{hatemRehabilitationMotorFunction2016,alawiehFactorsAffectingPoststroke2018}.

While there is no consensus on the best way to address this lack of progress, there exists substantial evidence that both high therapy dosage~\cite{roseRoadForwardUpperextremity2021, lohseMoreBetterUsing2014, gassertRehabilitationRobotsTreatment2018,salvalaggioPredictiveFactorsDose2024,gauthierDoseResponseUpper2024} \emph{and} high quality of practiced movements~\cite{gauthierDoseResponseUpper2024, langDoseResponseTaskspecific2016, pilaImpactDoseCombined2022a} are critical to promoting neuromotor recovery. 
Repeated execution of maladaptive compensatory patterns --- such as flexion synergy expression and trunk tilt --- may be reinforced through use-dependent plasticity, even when task success improves, thereby entrenching pathological coordination and constraining long-term recovery~\cite{clarkMergingHealthyMotor2010, ellis2009progressive, seoAlterationsMotorModules2022a}. 

To develop rehabilitation protocols that promote truly healthy engagement
requires not only quantifying whether or not a user’s motion is pathological, but determining how the system itself, including prescribed tasks, visual feedback, and robot interventions, should be adjusted to impel the user toward healthier coordination strategies rather than compensatory solutions. This task is further complicated by the intrinsic ``motor abundance''~\cite{latashBlissNotProblem2012} of the human neuromuscular system, wherein multiple distinct coordination patterns can achieve equivalent task outcomes, thereby rendering the precise definition and quantification of ``healthy'' motor behavior fundamentally difficult \cite{latashBlissNotProblem2012,valero2016fundamentals}.
This complex interplay is illustrated in Figure~\ref{fig:sum} in the context of our developed OpenRobotRehab rehabilitation platform~\cite{anandextensible2025}, in which goal trajectories are displayed to the user, who then executes motor commands to induce muscle forces, whose combination results in end-effector forces that are then translated to displayed avatar motions. 

In this work, we investigate the key questions noted in Figure~\ref{fig:sum}, toward ultimately designing robot control systems and  associated gaming frameworks that appropriately intervene to promote healthy neuromotor engagement. For these preliminary analyses, we leverage the OpenRobotRehab~1.0 multimodal data set, consisting of 8-channel surface electromyography (sEMG), 6D force torque, and game performance data collected from 13~healthy and 2~post-stroke participants during performance of various 6D isometric trajectory tracking tasks~\cite{anandextensible2025}.
To contextualize our findings on therapy-relevant differences between healthy and pathological behaviors, we examine these questions ``against'' the illustrated gaming feedback loop: first, investigating the influences of various game design decisions on user behavior for both healthy and post-stroke users; second, quantifying how healthy and pathological motor behaviors are reflected in end-effector force output and thus game performance; and third, establishing methods to discriminate between healthy and pathological muscle activation patterns. Our specific contributions include:
\begin{itemize}
\vspace{-0.25em}
    \item assessment of how oft-ignored aspects of trajectory tracking task specification (including unconstrained force directions and under-specified target following instructions) substantially impact user behavior; 
    \item quantification of differences in healthy and post-stroke game performance and associated exertion, illustrating clear pathology-associated disparities but also substantial heterogeneity in behavior of healthy participants; and
    \item a novel hidden Markov model (HMM)--based classification method (applied to sEMG data of the arm during cyclic task completion), which we show to be capable of discriminating between healthy and pathological neuromotor behavior
    when more standard synergy-based decompositions fail, and in some cases of exposing temporal features corresponding to observed compensatory motions.
\end{itemize}
Together, these insights and tools highlight important considerations for the design and evaluation of robot rehabilitation systems and gaming interfaces, and will inform our own expansions of the Figure~\ref{fig:sum} platform to enable effective and engaging robot-mediated therapy.

\vspace{-0.75em}
\section{Background and Related Work}
\label{sec:related}

In the field of rehabilitation robotics, multiple schools of thought have emerged regarding the primary drivers of effective therapy. A widely adopted perspective emphasizes that therapy dosage --- defined as the amount and intensity of practice --- is a key determinant of functional recovery~\cite{roseRoadForwardUpperextremity2021, lohseMoreBetterUsing2014, gassertRehabilitationRobotsTreatment2018,salvalaggioPredictiveFactorsDose2024,gauthierDoseResponseUpper2024}.
At the same time, a growing body of research cautions that optimizing systems for engagement, dosage, and task completion without accounting for motion quality may reinforce maladaptive motor strategies (e.g., the flexion synergy in post-stroke individuals) and compensatory motions (e.g., trunk lean, shoulder elevation). Randomized clinical trials have shown that simply increasing therapy dosage, in the absence of neurologically informed intervention, often yields diminishing returns in functional recovery \cite{gauthierDoseResponseUpper2024, langDoseResponseTaskspecific2016, pilaImpactDoseCombined2022a}. Advocates of this perspective emphasize the incorporation of neurophysiological principles into robotic system design to support reacquisition of healthy movement patterns, rather than merely optimizing controllers for task success.

However, integrating these principles into robotic systems remains a significant challenge. While certain motor disorders exhibit distinct movement abnormalities, defining what constitutes ``healthy'' motor behavior is complicated by substantial inter-individual variability in natural movement strategies \cite{latash2024useful,valero2016fundamentals}. 
To an extent, motion pathology is often observable in system users' force interaction with the robot end effector (with impaired users exhibiting, e.g., increased task error\cite{lodhaForceControlDegree2010,kangForceControlChronic2015} and changes in magnitude and variance characteristics~\cite{kangForceControlChronic2015, seoAlterationsMotorModules2022a}).
Prior studies have shown that post-stroke individuals demonstrate elevated force variability, impaired scaling of force amplitude, and reduced steadiness during sustained or cyclic contractions, reflecting deficits in motor unit recruitment, rate modulation, and intermuscular coordination~\cite{lodhaForceControlDegree2010,kangForceControlChronic2015}. Moreover, alterations in motor module structure have been directly linked to limitations in upper-extremity force control, suggesting that abnormal force dynamics at the endpoint are at least partially driven by disrupted patterns of coordinated muscle activation~\cite{seoAlterationsMotorModules2022a}.
At the same time, these force-based metrics alone offer a limited view of the underlying neuromotor processes that generate movement --- the disruption of which is the true cause of resulting motion pathology. Endpoint force represents the net mechanical consequence of a high-dimensional neuromuscular control strategy; similar force trajectories can arise from distinct combinations of muscle activations, co-contraction levels, and compensatory recruitment patterns. As a result, force error or variability may reveal that performance is degraded, but not how neuromotor coordination is altered, nor whether task success is achieved through maladaptive strategies (e.g., excessive co-contraction or abnormal synergy recruitment) that may reinforce pathological patterns over repeated practice~\cite{seoAlterationsMotorModules2022a}.

Two broad categories of approaches have emerged to model these neuromotor processes, and to discriminate between healthy and pathological behavior at the neuromuscular level. First, empirical, data-driven methods --- perhaps the most common of which is muscle synergy analysis~\cite{bizzi2013neural,park2023relevance,alessandroMuscleSynergiesNeuroscience2013}, in which complex muscle activations are decomposed into a reduced set of co-activated muscle modules --- aim to identify low-dimensional representations of neuromotor behavior that reflect the differences between normative and pathological motor strategies from time series muscle activation data. Such methods have been shown to enable characterization of pathology-specific deviations in coordination patterns~\cite{ortega-auriolRoleMuscleSynergies2025,dipietroChangingMotorSynergies2007, facciorussoMuscleSynergiesUpper2024}, and, newly, have been used to formulate therapeutic visual feedback strategies aimed at restoring more normative synergy profiles~\cite{berger2024myoelectric}. At the same time, while some limited pathology-associated patterns have been consistently replicated (e.g., the tendency of impaired individuals to exhibit fewer synergies than their healthy counterparts when accomplishing a given task~\cite{ortega-auriolRoleMuscleSynergies2025,davellaControlFastReachingMovements2006,clarkMergingHealthyMotor2010}), it remains contentious whether identified synergies genuinely reflect the organization of the human nervous system or instead arise as convenient basis functions of the observed experimental data~\cite{derugyAreMuscleSynergies2013,treschCaseMuscleSynergies2009}. This ambiguity limits the extent to which specific synergy patterns can be meaningfully classified as ``healthy'' or ``unhealthy'', and correspondingly constrains their utility for informing neuromotor therapies. Moreover, synergy-based models offer limited interpretability beyond coarse discrimination between impaired and unimpaired behavior --- often reduced to differences in the ideal number of synergies --- and therefore do not provide clear or actionable therapeutic pathways for improvement.

Second, neuromusculoskeletal (NMSK) simulation--based methods aim to infer detailed joint kinetics, muscle forces, and neuromotor activation dynamics from peripheral biosensor data to generate physiological models of the complete human NMSK system that exhibit pathological motor behaviors of interest. The appeal of such methods for characterizing motor behavior is straightforward: biomarkers of pathology (e.g., abnormal co-contraction, trunk tilt, flexion synergy expression) can be directly extracted from the simulation and compared across individuals. In practice, however, even state-of-the-art simulators (e.g., OpenSim~\cite{delp2007opensim}, MyoSuite~\cite{caggiano2022myosuite}) heavily rely on static optimization procedures and normative datasets derived primarily from cadaver specimens, healthy adults and/or limited subsets of target population groups~\cite{harringtonMusculoskeletalModelingMovement2024}, hindering accurate representation of pathology, and have achieved only limited adoption in upper-limb rehabilitation contexts as they are traditionally targeted towards lower-limb modeling~\cite{reinboltSimulationHumanMovement2011,sethOpenSimSimulatingMusculoskeletal2018}.

To date, these methods of identifying healthy and pathological motor behavior, whether decomposition- or simulation-based, have informed few robot-mediated rehabilitation systems, the majority of which remain reliant on control strategies that incentivize user engagement of any kind regardless of motion quality (e.g., assistance-as-needed (AAN)~\cite{blank2014current}, intent-triggered~\cite{blank2014current}, and dynamic difficulty adjustment (DDA)~\cite{pezzera2020} frameworks). While a few isolated breakthroughs have improved robots' ability to incentivize healthy neuromotor practice --- most notably, the insight that performing gravity compensation to reduce shoulder loading diminishes individuals' expression of the flexion synergy~\cite{ellis2009progressive,ellis2018progressive} --- systems that comprehensively assess users' individual deficits and prescribe corresponding therapy tasks and robot interventions to impel healthy practice remain elusive. 
Constructing such systems will require not only improved methods of discriminating healthy and pathological neuromotor behavior and better methods to interpret these findings, but a holistic understanding of how that behavior is shaped by specific physical and virtual interface properties --- including task instruction, visual feedback, and robot control algorithms --- of a given rehabilitation platform. 

This paper presents the first steps toward this type of holistic understanding of the complex interplay between human motor behavior, task specification, and robot interventions, and thus toward the development of rehabilitation robot platforms that provably impel healthy neuromotor practice. In previous work~\cite{anandextensible2025}, we constructed a modular, extensible end-effector rehabilitation robot gaming platform enabling the simultaneous measurement of muscle activation at the arm (via sEMG) and force torque data at the hand (via load cell), during performance of isometric trajectory tracking tasks in the full 6D space (3D positional, 3D rotational) of possible hand exertions, 
and collected this sensor data for a pilot cohort of 13 healthy and 2 post-stroke individuals performing 8 such tasks in 2 pose conditions. In this work, we employ that data set to investigate first, the most notable ways in which our platform's interface impacted user behavior;
second, the extent to which healthy and pathological behavior could be identified from end effector forces alone; and third, the extent to which standard synergy-based analysis tools --- as well as our own novel HMM-based decomposition method --- permitted healthy and pathological motion discrimination.

While the analyses we perform are inherently specific to our system --- and will be used to inform our own future platform augmentations, as outlined in section~\ref{sec:conc} --- both the implications we uncover for interface design and the methods we identify to characterize healthy and pathological neuromotor behavior highlight important considerations for the development of all rehabilitation robot systems.

\vspace{-0.75em}
\section{Methodology}
\label{sec:methods}
To accomplish the goals above --- namely, to first characterize the most salient impacts of our rehabilitation platform's interface design on user behavior, then develop strategies to discriminate between healthy and pathological neuromotor practice --- we analyze the data set detailed in section~\ref{methods:dataset}, preprocessed as noted in section~\ref{methods:preprocessing}, using the tools described in sections~\ref{methods:influence}, \ref{methods:force} and \ref{methods:decomposition}.

The study protocol employed in this work was approved by the University of Utah Institutional Review Board under Protocol IRB\_00154923 (approved 27 August 2022).

\vspace{-0.75em}
\subsection{Data Set}

All analyses were performed on the OpenRobotRehab 1.0 data set, released open-source on SimTK ({\tt \href{https://simtk.org/projects/openrobotrehab}{simtk.org/projects/openrobotrehab}}) alongside our previous work~\cite{anandextensible2025} documenting initial rehabilitation platform construction. This data set consists of time series surface electromyography (sEMG), force torque, and game performance data from healthy and post-stroke individuals performing isometric upper-limb trajectory tracking tasks on an end-effector robot. Full collection protocol information is available in the associated publication~\cite{anandextensible2025}, with details relevant to this study noted below.
\label{methods:dataset}

\subsubsection{Participant demographics}
Data were collected from 13 healthy individuals and 2 stroke survivors, for a total of 15 participants. The 13 healthy participants included 7 male and 6 female, and 9 right-handed and 4 left-handed individuals, of ages $29.5 \pm 14.0$ (mean $\pm$ standard deviation, min 20, max 70). Of the two stroke survivor participants, both in the chronic phase of pediatric stroke, the first was female, age 24, right-side hemiparetic, and reported being right-handed prior to her stroke. The second was male, age 34, left-side hemiparetic, and did not report handedness prior to neurological injury. Both stroke survivors exhibited mild spasticity (2 and 1, respectively, at the hand on the Modified Ashworth Scale \cite{bohannonInterraterReliabilityModified1987}) and mild decreases in muscle strength (4/4 and 4+/4+, respectively, for flexion and extension of the wrist on the Manual Muscle Testing scale \cite{ManualMuscleTesting}).\footnote{For additional demographic data, broken down by participant, see the full open-source data release.}

\subsubsection{Rehabilitation platform}
\begin{figure*}[tb]
    \centering
    \includegraphics[width=0.9\linewidth]{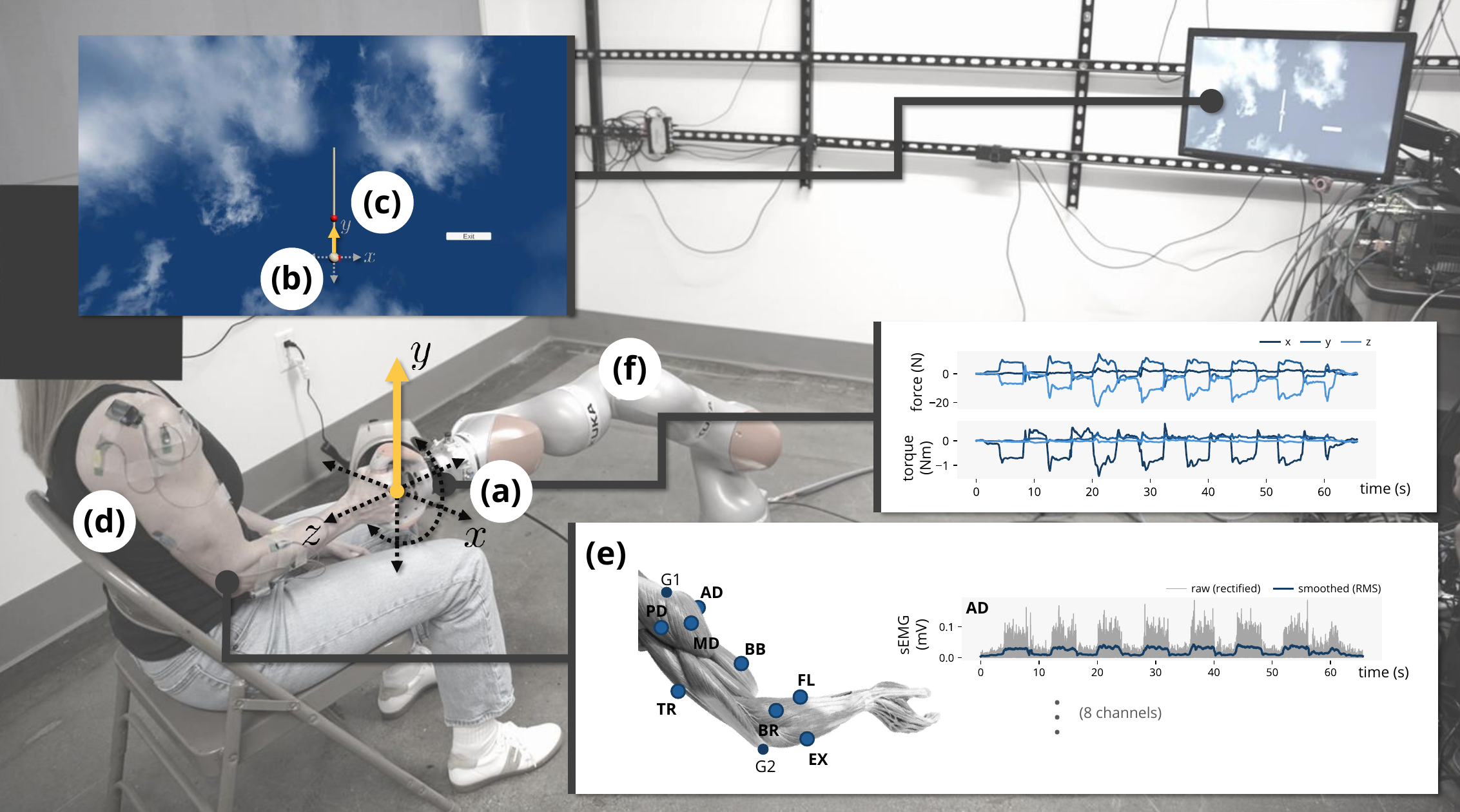}
    \caption{Motor rehabilitation platform enabling measurement of muscle engagement and end effector forces during trajectory tracking tasks. Users exert forces and torques on a SensONE 6-axis load cell (a) (Bota Systems AG, Zürich, Switzerland) through the attached handle, which are then mapped to $x$--$y$ coordinates of on-screen avatar (b) to allow users to follow red target ball (c) through different trajectories within a custom gamified rehabilitation environment developed in Unity (Unity Software Inc., San Francisco, CA, USA). During each trajectory tracking task, surface electromyography (sEMG) electrodes (d) --- two Trigno Quattro 4-channel sensor motes, controlled through a Trigno Base Station (Delsys, Inc., Natick, MA, USA) --- placed on key muscles of the arm (e) record muscle activations. The system currently supports isometric rehabilitation tasks at arbitrary poses --- the LBR iiwa 14 R820 7-degree-of-freedom cobot (f) (KUKA AG, Augsburg, Germany) remains static --- but will be expanded in the future to support a variety of robot controllers. Surface EMG electrode placements: anterior deltoid (AD), middle deltoid (MD), posterior deltoid (PD), and biceps brachii (BB), grounded at the acromion (G1); triceps brachii (long head, TR), brachioradialis (BR), wrist flexors (FL), and wrist extensors (EX), grounded at the olecranon (G2). Reprinted from~\cite{anandextensible2025}.}
     \label{fig:hardware}
     \vspace{-1em}
\end{figure*}

Participants completed trajectory tracking tasks on the rehabilitation gaming platform depicted in Figure~\ref{fig:hardware} while sensorized as noted.

\begin{figure*}[tb]
    \centering
        \includegraphics[width=0.95\linewidth]{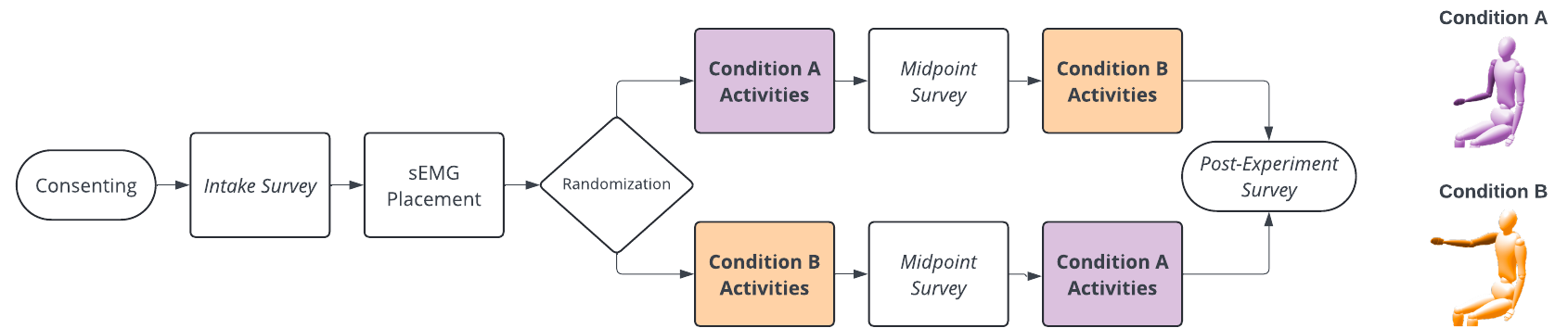}
    \caption{Experimental flow during collection of pilot data set. Participants were consented and surveyed, then completed trajectory tracking tasks at two ADL-inspired poses --- Conditions A (shoulder adducted, elbow flexed $\sim 90\degree$, forearm horizontal) and B (shoulder slightly flexed above horizontal with near full elbow extension) --- in randomized order before providing final survey feedback. Survey data
    are not examined in this work. Reprinted from~\cite{anandextensible2025}.}
    \label{fig:exptflow}
    \vspace{-1em}
\end{figure*}

\subsubsection{Collection protocol}
Participants followed the experimental flow shown in Figure~\ref{fig:exptflow}. Once consented, and after performing an intake survey and being sensorized with sEMG as shown in Figure~\ref{fig:hardware}, participants completed the 8 trajectory tracking tasks listed in Table~\ref{tab:trajtrack} for the specified number of repetitions in each of 2 pose conditions in randomized order while time series sEMG, force torque, and game performance data were collected. Unlimited practice and verbal readiness checks were permitted prior to each task recording, and mandatory 2-minute rests separated the first 3, second 3, and final 2 tasks, with additional breaks provided as needed, to combat fatigue. Participants then provided final survey feedback and exited the study.

\subsection{Pre-Processing}
\vspace{-0.25em}
\label{methods:preprocessing}
For each trial, force signals were temporally trimmed to match each task’s duration (as measured via game start and end time) and baseline-corrected by removing steady-state offsets via linear regression on each $(x, y, z)$ axis (calculated per participant and condition). Surface EMG signals (collected at 1~kHz) were similarly segmented, band-pass filtered (30--450~Hz, 4th-order Butterworth), rectified, and smoothed using a 400-sample RMS window, then normalized per muscle using each participant’s maximum activation across all trials \cite{konradABCEMGPractical2005a}.
\vspace{-0.75em}
\subsection{Analysis tools: Influence of game design and target-tracking strategies on user behavior}
\label{methods:influence}

While every design decision 
--- from the input--output display mapping decisions documented in Table~\ref{tab:trajtrack} to the timing of visual cues
--- potentially impacts user behavior, we focus our attention on the following phenomena, guided by our qualitative observations during data collection.

\subsubsection{Prevalence of ``non-productive'' force generation}
\label{methods:non-productive}
Accomplishing each trajectory tracking task requires ``productive'' effort in one or more dimensions: e.g., the $x$-axis task requires horizontal motion, and the circle tasks require tracing a particular path in $x-y$ space. Simultaneously, all tasks leave one or more force dimensions --- which we term ``non-productive'' --- unspecified: e.g., $y$- and $z$-axis motions have no impact on $x$-axis task performance, and $z$-axis motions no impact on circle tasks.\footnote{Note that the same is true of productive and non-productive torques, though they are not analyzed in this work.} 
We examine the characteristics of these non-productive forces, both temporally and in aggregate, across tasks, impairment levels, and pose conditions. Specifically, we employ the same force RMSE, force impulse, and average and peak force output measures described below in section~\ref{methods:force}, applied only to non-productive force components.

\subsubsection{Heterogeneity of user target tracking strategies}
\label{methods:target-tracking}

\begin{figure}[tb]
    \centering
    \includegraphics[width=0.7\linewidth]{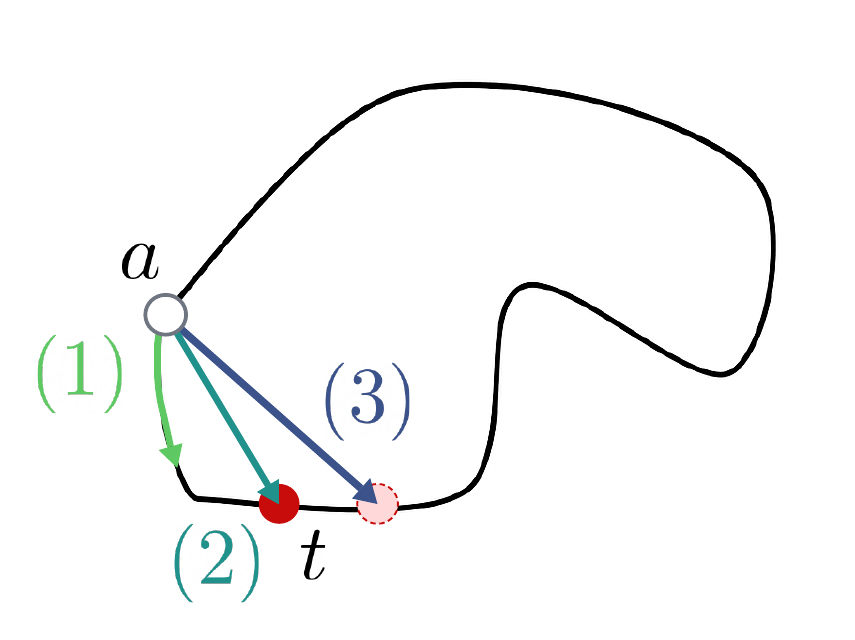}
    \caption{
        Illustration of three distinct observed target tracking behaviors, depicted during the spline 2 trajectory tracking task.
        Participants were instructed to follow target ball $t$ with avatar ball $a$ as it traversed the specified trajectory. Observed strategies included (1)~pursuit of the target along the trajectory, (2)~shortest-distance pursuit to $t$'s current position, and (3)~movement toward an anticipated future position of $t$.
    }
    \label{fig:gameStrategies}
    \vspace{-1.5em}
\end{figure}

While trajectory tracking tasks were in some aspects tightly specified, 
observations during data collection revealed multiple tracking strategies used by participants (the most salient of which are illustrated in Figure~\ref{fig:gameStrategies}), each of which was consistent with task specifications but required different underlying neuromotor behavior. To corroborate these observations, we examine time series trajectories of the target and player avatar for evidence of these behaviors and remark on their prevalence and implications for interface design.

\begin{table}%
    \centering
    \begin{threeparttable}
    \caption{Trajectory Tracking Tasks \& Mappings}
    \label{tab:trajtrack}
    \def\arraystretch{1.1}%
            \begin{tabular}{c|c|c|c}
                    Goal Trajectory & Input & Output (Game) & Repetitions  \\
                    \hline
                    $x$-axis & $x$-axis force & $x$ coordinate & 7 \\
                    $y$-axis & $y$-axis force & $y$ coordinate & 7 \\
                    $z$-axis & $z$-axis force & $y$ coordinate & 7 \\
                    torque & $z$-axis torque & $x$ coordinate & 5 \\
                    circle (CW$^{1}$) & $(x,y)$-axis force & $(x,y)$ coordinate & 3 \\
                    circle (CCW$^{2}$) & $(x,y)$-axis force & $(x,y)$ coordinate & 3 \\
                    spline 1 (CCW$^{2}$) & $(x,y)$-axis force & $(x,y)$ coordinate & 3 \\
                    spline 2 (CW$^{1}$) & $(x,y)$-axis force & $(x,y)$ coordinate & 3 \\
                    \hline 
            \end{tabular}
            \begin{tablenotes}
            \scriptsize
            \item Screenshots of each trajectory are included with the data release \cite{anandextensible2025}. \item $^1$clockwise  $^2$counter-clockwise
            \end{tablenotes}
    \end{threeparttable}
    \vspace{-1.5em}
\end{table}

\subsection{Analysis tools: Force-based discrimination of healthy
and pathological  behavior}
\label{methods:force}

To determine the extent to which we can discriminate healthy and pathological motor behaviors from the force and game data intrinsically available during game play, we assess users' performance and exertions in the ``productive'' dimension(s) for each task via the following literature-grounded metrics. For tasks with a single associated productive force direction, metrics are computed on the time series force in that dimension; in tasks with multiple productive dimensions, metrics are computed on the time series L2 norm of the force in these productive dimensions.
Note that we restrict our analyses to forces and thus omit the torque task from analysis; the aggregate metrics below are reported on the mean and standard deviation of the remaining 7 tasks.

\subsubsection{Force RMSE}
\label{methods:rmse}

Prior work has shown that stroke survivors frequently exhibit increased task error, elevated variability in force output, and prolonged delays in the initiation and coordination of muscular forces \cite{kangForceControlChronic2015, seoAlterationsMotorModules2022a}.
To quantify such errors, we perform a root mean square error (RMSE) force analysis by comparing each participant’s force profile along the task-relevant (i.e., productive) axis or axes to idealized force trajectories associated with optimal task performance.\footnote{These idealized trajectories are calculated by inverting the equation used to translate forces to in-game cursor motion; see appendix~I for details. To address the mismatch in sampling frequencies between the game and the force sensor, a nearest-neighbor matching approach was applied with a temporal threshold of 1 ms.}

\subsubsection{Force impulse}

Rectified force impulse --- derived by integrating the absolute force magnitude over time --- provides an estimate of the mechanical work demand placed on the musculoskeletal system, which is known to correlate with metabolic cost under isometric conditions \cite{sihMetabolicCostForce2003}. We examine this metric in aggregate across healthy and post-stroke cohorts.

\subsubsection{Average and peak force output}
To quantify differences in force production patterns contributing to observed force RMSE, we analyze both the root mean square (RMS) average and peak values of the rectified force signal in aggregate across healthy and post-stroke cohorts. 

 \subsubsection{Statistical tests}

We assess group differences in the above metrics between healthy and post-stroke cohorts using the Mann–Whitney U test, a non-parametric test for independent samples, with significance set at $p<0.05$,
generating plots using the Seaborn data visualization library \cite{Waskom2021}.
\vspace{-0.5em}
\subsection{Analysis tools: Decomposition-based discrimination of healthy and pathological patterns of neuromuscular activation}
\label{methods:decomposition}

To truly characterize healthy and pathological motor behavior --- and, ultimately, to modify it via robot intervention --- requires quantifying not only output forces (which may be task-performant but rely on maladaptive compensatory behaviors), but the underlying neuromuscular dynamics that result in those forces. Toward such insights, we construct and evaluate the following models.

\subsubsection{Synergy decomposition}

Although its grounding in real neurophysiological behavior remains contentious ~\cite{derugyAreMuscleSynergies2013,treschCaseMuscleSynergies2009}, muscle synergy decomposition --- in which EMG data are decomposed using nonnegative matrix factorization (NMF) \cite{funato2022muscle, valero2016fundamentals} or other factorization methods --- is perhaps the most common dimensionality reduction technique used to assess neuromotor behavior. The synergy abstraction admits several avenues of analysis to quantify pathology. First, prior work suggests that a lower observed number of synergies may indicate motor impairment ~\cite{ortega-auriolRoleMuscleSynergies2025,dipietroChangingMotorSynergies2007, facciorussoMuscleSynergiesUpper2024}. Second, users' synergy decompositions can be compared to identify clusters of individuals with similar (healthy or pathological) motor coordination patterns ~\cite{seoAlterationsMotorModules2022a,rohAlterationsUpperLimb2013,rohEvidenceAlteredUpper2015}. 

To enable both of these lines of analysis, we decompose each 8-channel sEMG time series (for all participants, tasks, and pose conditions) using NMF, resulting in trial-specific sets of synergies. Next, we assess the optimal synergy count (OSC) required to adequately reconstruct the recorded muscle activity by computing the variance accounted for (VAF) by candidate decompositions of 1--8 synergies \cite{rohAlterationsUpperLimb2013,rohEvidenceAlteredUpper2015}. Consistent with established work\cite{rohAlterationsUpperLimb2013,rohEvidenceAlteredUpper2015}, we define the OSC as the smallest number of synergies for which the VAF exceeds 0.90, and the incremental increase in VAF from adding an additional synergy is less than 0.03.
Finally, to detect cohorts with similar motor behavior, we perform $k$-means clustering ~\cite{scanoMuscleSynergiesBasedCharacterization2017} on these optimal synergy decompositions. Details of these analysis procedures are documented in appendix~II.

\subsubsection{Hidden Markov model (HMM)--based decomposition}
\label{methods:HMM}

While synergy decomposition methods dominate approaches to empirical evaluation of neuromotor dynamics, successful synergy-based discrimination of healthy and pathological behavior has to date been largely restricted to stereotyped, repeated motions under tightly controlled conditions ~\cite{seoAlterationsMotorModules2022a, rohAlterationsUpperLimb2013,rohEvidenceAlteredUpper2015, davellaControlFastReachingMovements2006} better reflected in lower limb motions like the gait cycle than the natural unconstrained motions of the arm that we ultimately hope to rehabilitate. Noting this limitation --- as well as our own lack of success, documented in section~\ref{results:synergies}, in employing synergy-based decompositions to identify pathology --- we propose hidden Markov models (HMMs) as an alternative decomposition of motor behavior.

HMMs~\cite{rabinerTutorialHiddenMarkov1989} are widely used to identify underlying states in time series data. This capability has been exploited in a (sparse) number of neuromechanical studies to infer intent~\cite{7839943}, classify actions being performed~\cite{WEN2021102592}, and characterize latent relationships between muscle activations \cite{4520143} from sEMG data. In this work, we identify an additional use for this decomposition: segmenting temporal sEMG data into underlying motor actions, implicitly identifying behavior that fails to conform to the cyclic motion prescribed by our task trajectories, which we expect to indicate underlying neuromotor pathology.

More explicitly: we note that successful, healthy completion of the prescribed trajectory tracking tasks --- each of which involves multiple repetitions of the prescribed motion, as noted in Table~\ref{tab:trajtrack} --- requires participants to alternate between two or more subtasks (e.g., left and right motions on the $x$-axis task), and that failure to display two corresponding underlying modes of neuromuscular activation
may reflect maladaptive neuromotor strategies. We hypothesize that these subtask states will be reflected in the action sequence (i.e., Viterbi path \cite{1450960}) identified by an HMM appropriately fitted to the data. This theory suggests the following relationship between the HMM-predicted action sequence on a given trajectory tracking task and a participant's neuromotor health: the more the predicted action sequence --- i.e., the predicted sequence of neuromuscular engagement modes --- diverges from the subtask sequence prescribed by the target trajectory,
the more pathological their neuromotor dynamics.

To test this hypothesis --- i.e., to validate this relationship --- for each trial for each participant, we fit an HMM using standard expectation maximization methods~\cite{rabinerTutorialHiddenMarkov1989} to the 8-channel sEMG time series corresponding to each of the 4 single-axis trajectory tracking tasks noted in Table~\ref{tab:trajtrack}, assuming 2 discrete states corresponding to distinct directional actions.\footnote{We constrain our analysis to single-axis tasks, as they can most readily be decomposed into distinct directional subtasks; the behavior of HMM-based decompositions on tasks for which subtask decompositions are less obvious is a compelling area for future research.} We then calculate the HMM \emph{subtask classification error} --- our defined metric for action sequence inconsistency, and thus neuromotor impairment --- as the fraction of time points for which the HMM predicted an action inconsistent with the subtask specified by the target trajectory.\footnote{The precise formulation of this error calculation is described in appendix~III.} HMM fitting is probabilistic, and we report the mean and variance of this metric as calculated over 25 independent HMMs fit to each trial.

\vspace{-1.5em}
\section{Results and Discussion}
\label{sec:results}

Leveraging the data set and methods in the preceding section, we report the following findings and discuss their implications for the design of end-effector rehabilitation robot platforms and the characterization of healthy and pathological neuromotor behaviors when interacting with such systems.

\vspace{-1.25em}
\subsection{Influence of game design and target tracking strategies on user behavior}
\label{results:implications}

Employing the analysis tools in section~\ref{methods:influence}, we note the following findings on the impacts of interface design and provided instructions on observed behavior and task performance, across both healthy and post-stroke participants.

\begin{figure}[tb]
    \centering
    \includegraphics[width=\linewidth]{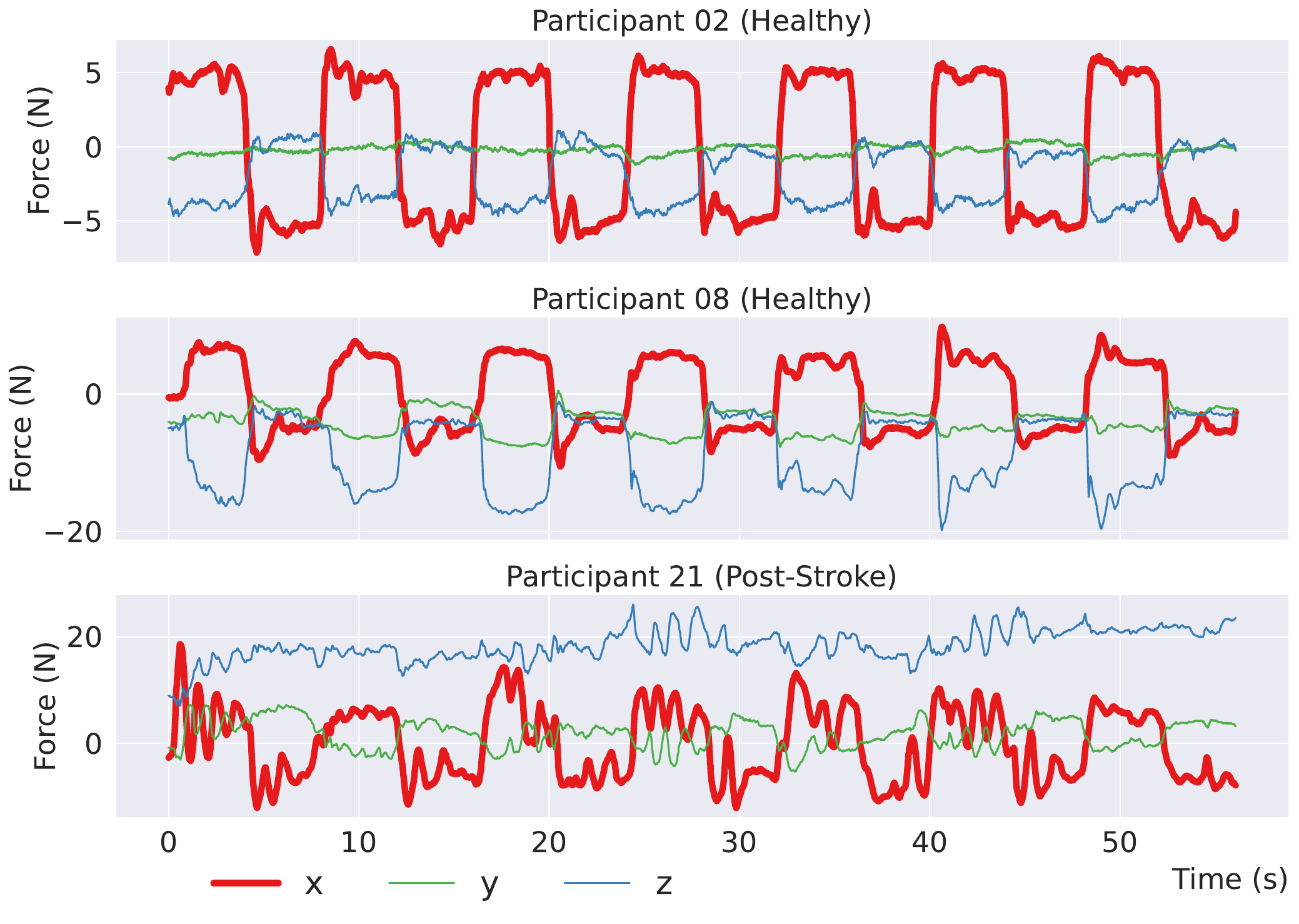}
    \caption{
        Exemplar time series force profiles of two healthy participants (02, \emph{top} and 08, \emph{middle}) and one post-stroke participant (21, \emph{bottom}) performing the $x$-axis trajectory tracking task in pose Condition A. All participants exhibit not only ``productive'' force exertions ($x$, \emph{red}) contributing to task completion, but substantial ``non-productive'' output in orthogonal dimensions ($y$, \emph{green} and $z$, \emph{blue}). The magnitude and temporal characteristics of these non-productive exertions vary within both healthy and post-stroke populations, illustrating the importance of careful specification when prescribing movement tasks. The post-stroke participant also exhibits markedly altered force production, with sustained off-axis force generation and reduced efficiency in producing task-relevant forces, reflecting underlying pathology.
    }
    \label{fig:force-comparison}
    \vspace{-0.5em}
\end{figure}

\begin{figure}[tb]
    \centering
    \includegraphics[width=\linewidth]{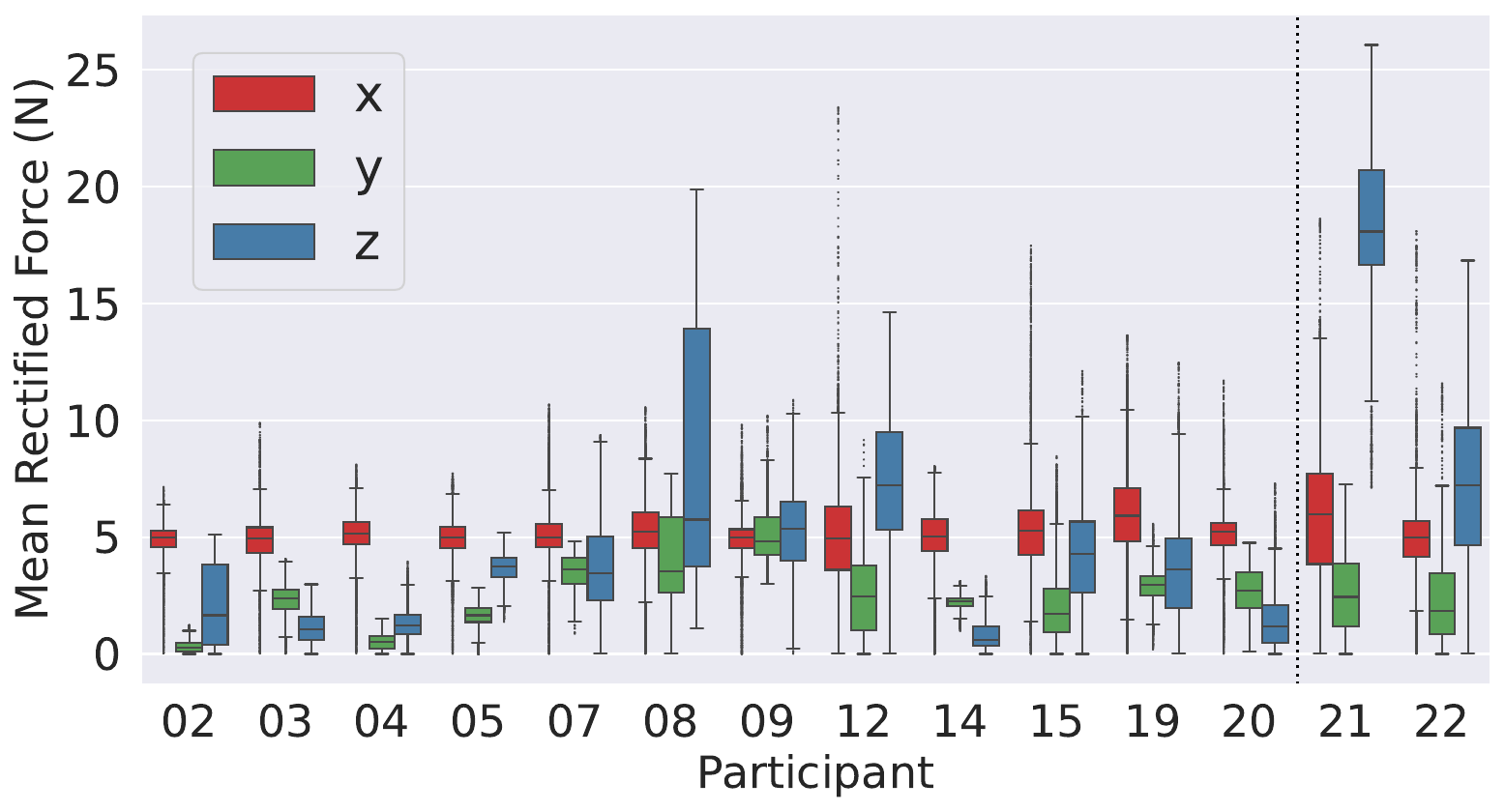}
    \caption{
        Aggregate force output --- measured as the mean and variance of each rectified force time series --- for all 3 force dimensions during completion of the $x$-axis tracking task in pose Condition A for both healthy (02--20, \emph{left}) and post-stroke (21--22, \emph{right}) participants. While productive $x$-axis (\emph{red}) force magnitudes are largely consistent across individuals, post-stroke participants exhibit higher variance and elevated levels of off-axis forces ($y$, \emph{green} and $z$, \emph{blue}), indicating increased non-productive effort. Healthy participant behavior is, however, heterogeneous, with some healthy individuals exerting substantial non-productive forces (e.g., 08, 12), and others (e.g., 04, 14) exhibiting only minor off-axis force production and thus improved task efficiency. In this and other tasks, most individuals exhibited highest and most variable non-productive exertions in the $z$-axis (i.e., in/out). Box plots show the median (50th percentile), interquartile range (25th--75th percentiles), whiskers extending to 1.5$\times$IQR, and outliers beyond this range.
    }
    \label{fig:force-summary}
    \vspace{-1.0em}
\end{figure}

\subsubsection{Task constraints --- or lack thereof --- strongly influence the magnitude of non-productive forces}

We observe substantial non-productive force components (as formally defined in section~\ref{methods:non-productive}), in both healthy and impaired participant groups across all pose conditions and tasks, revealing that a considerable proportion of users' applied force did not contribute directly to task completion. Figure~\ref{fig:force-comparison} illustrates this phenomenon as observed in 3 representative participants' completion of a single exemplar task, and Figure~\ref{fig:force-summary} shows aggregate force production over time across all participants for this task. These figures illustrate that while post-stroke participants exhibit generally higher and more variable non-productive exertions, healthy behavior is also heterogeneous, underscoring the extent to which individuals will behave differently within under-specified domains, 
as influenced by the specifics of how forces are transformed into the game environment and displayed on the screen, including how 3D forces are mapped to the 2D game display and which dimensions are disregarded.

One consistent finding
was that participants exhibited higher magnitude and larger variance of force along the $z$ (in/out) axis. This held for both the $z$-axis task, in which $z$ forces were productive (resulting in lower overall task performance due to the higher variance) and for the majority of the remaining tasks, in which $z$ forces were non-productive. This effect was especially pronounced in the post-stroke participants. We speculate that several factors likely contributed to this behavior. First,the $z$-dimension was more closely aligned with the axis of effort in both conditions, resulting in higher force measurements due to the geometry of the handle--load cell interface. Second, participants had less practice generating $z$-axis force, as it was targeted in only a single activity. Third, unlike $x$- and $y$-axis tasks, in which forces were mapped to the corresponding avatar motions on the 2D screen, the $z$-axis task was specified by mapping $z$-axis forces to the $y$-axis, likely reducing the intuitiveness of the mapping and thus reducing task performance. We speculate that these insights would likely be generalizable to all end-effector robot interactions.

Overall, these findings illustrate that non-productive forces are a nontrivial component of motor behavior --- even when tasks are specified relatively precisely --- and thus must be considered when analyzing underlying neuromuscular control strategies and when prescribing rehabilitative therapy tasks.

\subsubsection{User target tracking strategies are heterogeneous, resulting in varied force profiles reflecting varied underlying neuromotor behavior} 
\label{sec:motorStrategies}
\begin{figure}[tb]
    \centering
    \includegraphics[width=0.98\linewidth]{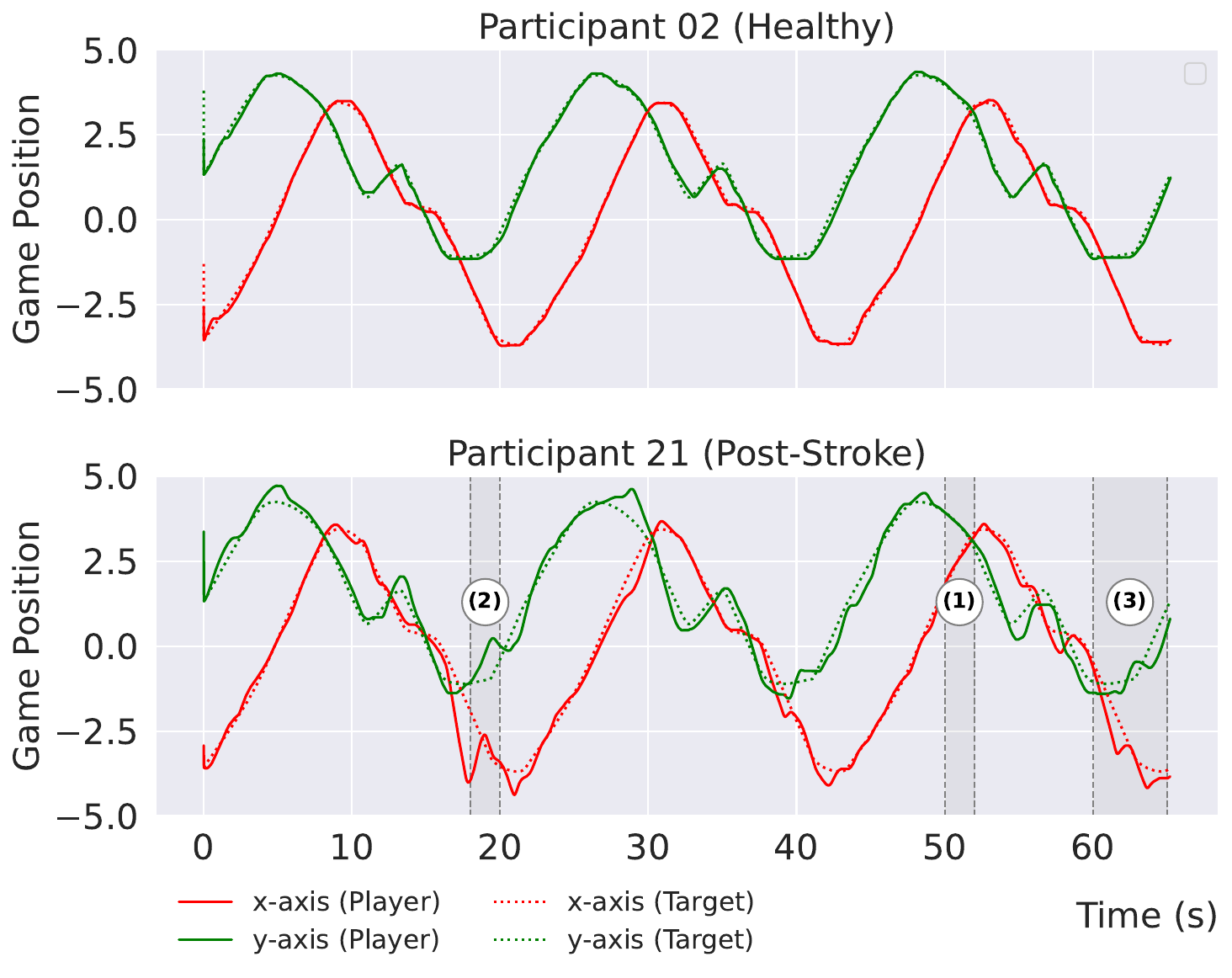}
    \caption{
    Exemplar time series data of both player avatar (\emph{solid}) and target ball (\emph{dashed}) $x-y$ position during completion of the spline 2 trajectory tracking task for two exemplar participants. While some participants, such as 02 (\emph{top}), performed sufficiently tight tracking that their trajectories were largely indistinguishable from that of the target ball, most participants (including 21, \emph{bottom}, when performing the displayed task), both healthy and post-stroke, exhibited multiple distinct target tracking behaviors to varied extents, which must be considered in analyses of task performance and comparisons of neuromotor behavior across individuals. Annotations (1), (2), and (3) indicate times at which each of the 3 pursuit strategies identified in Figure~\ref{fig:gameStrategies} can be observed within a single trial.
    }
    \label{fig:gameTrajectories}
    \vspace{-1.5em}
\end{figure}

Examination of time series trajectories of both the displayed target and the player-controlled avatar across all participants, tasks, and pose conditions reveals that the majority of users employ a heterogeneous mix of target tracking strategies, with most participants exhibiting all 3 strategies identified in Figure~\ref{fig:gameStrategies} at various times during various tasks,
as illustrated for an exemplar task in Figure~\ref{fig:gameTrajectories}.
These findings support the need to consider this heterogeneity when performing downstream analyses of user behavior (e.g., characterization of neuromotor activation patterns as in the analyses we perform below), and the importance of
carefully instructing participants in expected pursuit strategies when any of these observed behaviors will impact the efficacy of therapy protocols.

\vspace{-0.5em}

\vspace{-0.75em}
\subsection{Force-based discrimination of healthy and pathological behavior}
\label{results:force}
\vspace{-0.25em}
Comparing aggregate force RMSE, force impulse, and average and peak force output as defined in section~\ref{methods:force} across healthy and post-stroke cohorts for each pose condition, we gain the following insights on the differences between healthy and pathological force production at the end effector.

\begin{figure}[tb]
    \centering
    \begin{subfigure}[b]{0.2\textwidth}
        \centering
        \includegraphics[width=\textwidth]{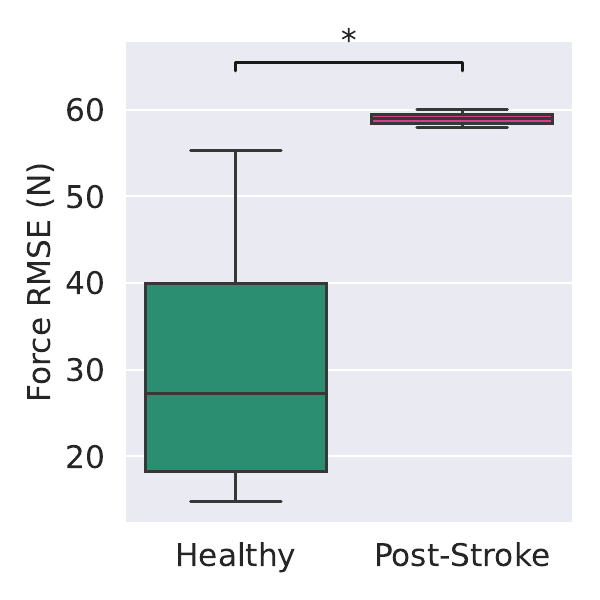}
        \caption{}
        \label{fig:forcesRMSE}
    \end{subfigure}
    \begin{subfigure}[b]{0.2\textwidth}
        \centering
        \includegraphics[width=\textwidth]{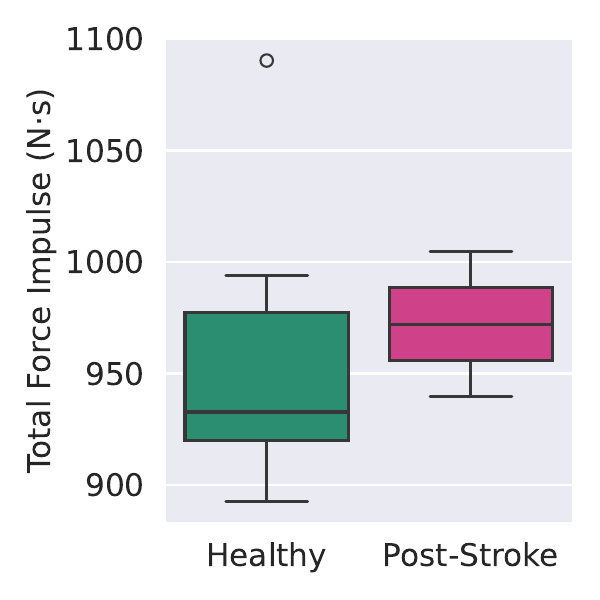}
        \caption{}
        \label{fig:forcesImpulses}
    \end{subfigure}
    \hfill
    \begin{subfigure}[b]{0.2\textwidth}
        \centering
        \includegraphics[width=\textwidth]{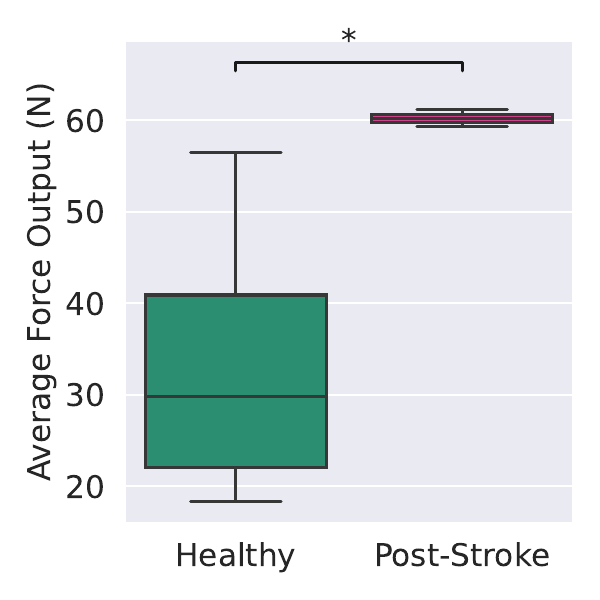}
        \caption{}
        \label{fig:forcesRMS}
    \end{subfigure}
    \begin{subfigure}[b]{0.2\textwidth}
        \centering
        \includegraphics[width=\textwidth]{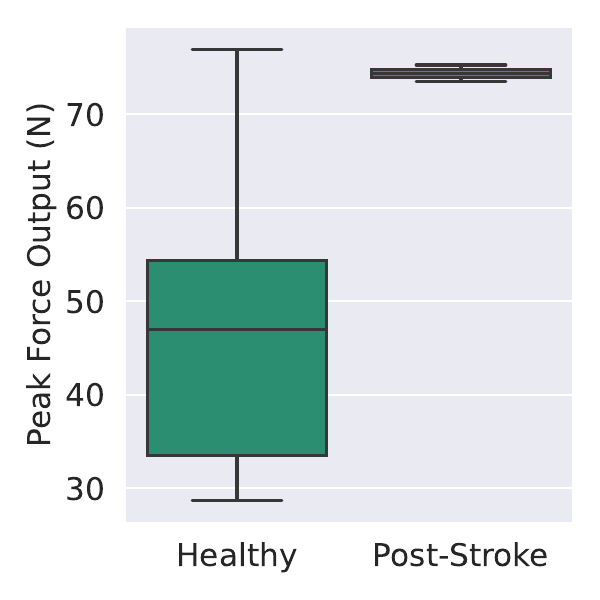}
        \caption{}
        \label{fig:forcesPeak}
    \end{subfigure}
    \caption{``Productive'' force RMSE (a), total force impulse (b), average force output (c), and peak force output (d), as calculated for each of the 7 force trajectory tracking tasks at pose Condition A, aggregated across all healthy (02--20, \emph{turquoise}) and post-stroke (21--22, \emph{pink}) participants. Post-stroke participants exhibit significantly higher force RMSE, consistent with existing literature, as well as significantly higher average force output. Total force impulse and peak force output appeared higher for impaired participants, but results were not statistically significant at $p=0.05$. These findings were consistent across both pose conditions, with only marginal differences between them. Box plots show the median (50th percentile), interquartile range (25th--75th percentiles), whiskers extending to 1.5$\times$IQR, and outliers beyond this range.}
    \label{fig:forcesAgg}
    \vspace{-1.0em}
\end{figure}

\begin{figure}[tb]
    \centering
        \includegraphics[width=0.95\linewidth]{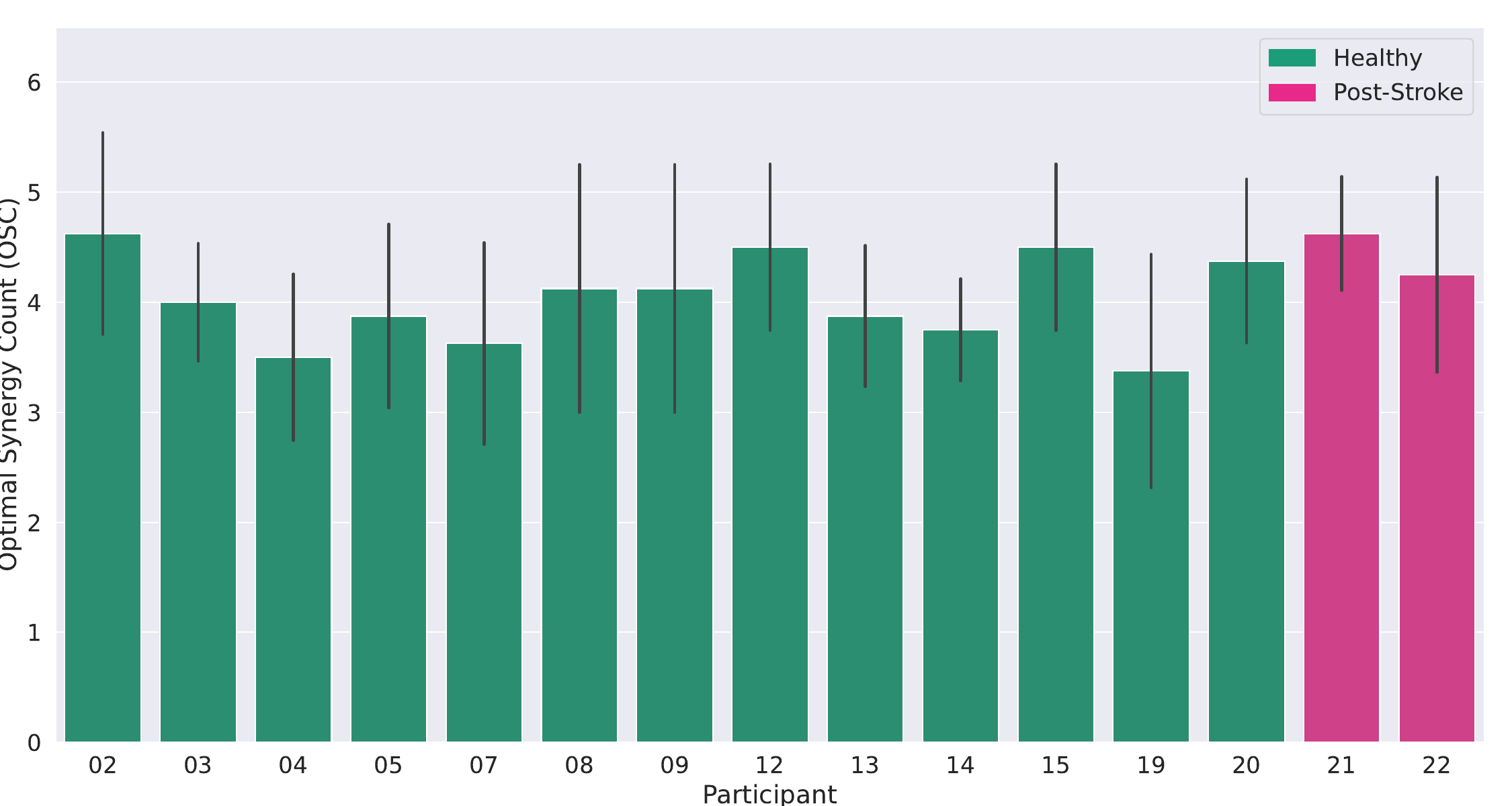}
    \caption{Optimal synergy count (OSC) reflecting neuromotor behavior complexity, aggregated across all tracking tasks in pose Condition A, for all participants. Both healthy (02--20, \emph{turquoise}) and post-stroke (21--22, \emph{pink}) participants exhibit comparable OSCs of 3--5 (standard deviation displayed as black lines), challenging previous findings~~\cite{ortega-auriolRoleMuscleSynergies2025} documenting an inverse relationship between OSC and impairment severity. These findings hold across both pose conditions with only marginal differences.
    }
    \label{fig:optimal-synergy}
    \vspace{-1.5em}
\end{figure}

\subsubsection{Pathology-associated effects are obvious on force production metrics even when aggregated across diverse 3D tasks}
\label{forces:pathology}

Force production metrics compared across healthy and post-stroke populations for Condition A are illustrated in Figure~\ref{fig:forcesAgg}, with corresponding plots for Condition B available in appendix~IV.
Post-stroke participants exhibited significantly higher levels of force RMSE with significant group differences observed in both Condition A (U = 0.00, p = 0.022) and Condition B (U = 0.00, p = 0.022); post-stroke means exceeded healthy means in both poses (A: 58.94 [95\% CI: 45.96--71.93] vs.\ 30.91 [21.80--40.03]; B: 59.02 [48.04--70.00] vs.\ 30.77 [21.58--39.97]), with complete separation ($r = 1.00$ in both conditions). They also exhibited significantly higher average force output, with significant differences observed in both Condition A (U = 0.00, p = 0.019) and Condition B (U = 0.00, p = 0.019); post-stroke participants exhibited higher means (A: 60.20 [48.42--71.99] vs.\ 33.71 [25.87--41.55]; B: 60.43 [49.83--71.03] vs.\ 33.82 [26.02--41.63]), again with complete separation ($r = 1.00$) aggregated across all tasks.

Post-stroke participants were also observed to exhibit higher levels of peak force output, but differences were not significant in either Condition A (U = 4.00, p = 0.171) or Condition B (U = 3.00, p = 0.114); post-stroke means were higher in both poses (A: 74.38 [62.95--85.81] vs.\ 48.37 [38.80--57.95]; B: 75.87 [56.46--95.27] vs.\ 48.57 [38.37--58.78]), with large effect sizes ($r = 0.69$, $0.77$). They also exhibited higher total force impulse aggregated across all tasks for each pose condition but no significant differences were exhibited either in Condition A (U = 6.00, p = 0.305) or Condition B (U = 3.00, p = 0.114); post-stroke means were modestly higher (A: 972.15 [559.75--1384.54] vs.\ 952.60 [921.79--983.40]; B: 1010.76 [472.05--1549.46] vs.\ 950.49 [927.09--973.89]), with moderate-to-large effect sizes ($r = 0.54$, $0.77$) and substantial variability. 

These results suggest that key characteristics of force production vary significantly between healthy and post-stroke populations, even when aggregated across diverse 3D trajectory tracking tasks, and are in accordance with prior work in this domain~\cite{lodhaForceControlDegree2010,kangForceControlChronic2015}. Notably, the qualitative pattern of results was consistent across pose conditions, indicating that pathology-associated differences in force production were preserved despite changes in upper-limb configuration. This is in contrast to the results of synergy-based decomposition presented in section~\ref{results:synergies}, highlighting the drawbacks of synergy analysis in capturing the underlying effects of pathology on neuromotor behavior.

\subsubsection{Healthy motor behavior is not a monolith}
In addition to the observed effects of pathology on force production noted in section \ref{forces:pathology}, both the relatively high variance of all four defined aggregate force metrics across healthy individuals and qualitative observations at the temporal level reveal substantial variability in productive force generation strategies even among healthy participants. 
These findings align with prior work documenting inter-individual differences in motor control, shaped by factors such as neuromotor redundancy, learning history, and individual movement preferences \cite{latashBlissNotProblem2012, valero2016fundamentals}. An illustration of the wide range of force production patterns can be observed in Figure~\ref{fig:force-comparison} depicting the heterogeneity of force production patterns in healthy individuals. This variability highlights that healthy motor behavior is not a monolith and complicates efforts to define a singular ``normal'' profile for motor execution.

\vspace{-1em}
\subsection{Decomposition-based discrimination of healthy and pathological patterns of neuromuscular activation}

We present the following findings on the efficacy of synergy- and HMM-based decompositions --- as defined in section~\ref{methods:decomposition} --- in enabling the classification (and, ultimately, modification) of pathological neuromotor behavior.

\subsubsection{Synergy decompositions do not expose differences between healthy and post-stroke participants}
\label{results:synergies}
As shown in Figure~\ref{fig:optimal-synergy}, healthy and post-stroke participants do not exhibit significantly different optimal synergy counts (OSCs), failing to replicate prior studies documenting reduced synergy count in impaired populations~~\cite{ortega-auriolRoleMuscleSynergies2025, clarkMergingHealthyMotor2010,davellaControlFastReachingMovements2006} and indicating that all participants employed neuromotor strategies of similar complexity. This negative finding may simply illustrate the limits of our data set --- which includes only two post-stroke participants with only mild impairment, performing only isometric exertions --- but also illustrates the limits of the OSC metric, which fails to capture the compensatory movement strategies investigators visually observed in even these mildly impaired participants during data collection (which included trunk tilt, flexion synergy expression, and other motion irregularities reflected in the force output differences documented in section~\ref{results:force}).

Clustering via $k$-means similarly failed to identify groupings of normative and impaired participants, and on inspection, optimal decompositions for both healthy and post-stroke participants largely contained qualitatively similar synergies. This may again reflect limitations of our data set --- and limitations of $k$-means clustering methodology, which is highly sensitive to hyperparameters such as the number of cluster centers --- but presents further evidence that synergy decompositions do not capture key aspects of neuromotor pathology, at least when evaluated on isometric exertions.

Synergy-based analyses are most often performed on highly constrained motor behaviors~\cite{seoAlterationsMotorModules2022a, rohAlterationsUpperLimb2013,rohEvidenceAlteredUpper2015, davellaControlFastReachingMovements2006}, so we considered one final explanation for these negative results: that the examined time series contained multiple distinct subtasks best described by separate synergy decompositions. To test this hypothesis, we segmented each single-axis trajectory tracking task into corresponding directional subtasks (as identified by the direction of the prescribed trajectory), then re-performed both OSC and $k$-means analyses separately on each subtask. These analyses yielded the same negative results above, further supporting the inadequacy of the synergy decomposition in capturing neuromotor pathology during isometric exercise.

\begin{figure}[tb]
    \centering
        \includegraphics[width=0.9\linewidth]{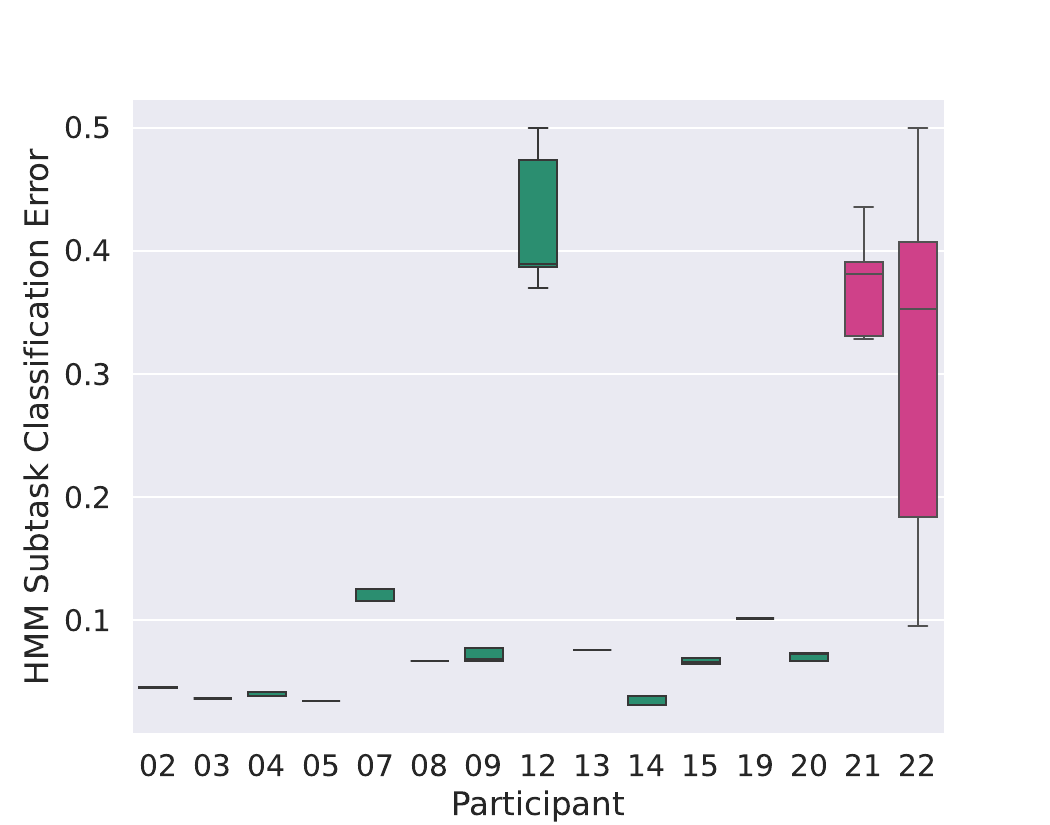}
    \caption{HMM subtask classification error --- defined in section~\ref{methods:HMM} as the fraction of time points for which HMM-predicted action sequences and target direction subtasks were not the same --- is higher in average magnitude and variance (as calculated across 25 independently trained HMMs for each trial) for post-stroke participants (21--22, \emph{pink}) for the $x$-axis trajectory tracking task
    as compared with most healthy participants (02--20, \emph{turquoise}), a finding that holds across the majority of examined tracking tasks in each pose condition (see appendix V), providing a metric for discriminating healthy and pathological neuromotor behavior. Some healthy participants (here, 12) exhibit behavior more consistent with post-stroke participants as quantified by this metric intermittently across tasks, illustrating the challenge of discriminating truly pathological motor behavior from uncommon but healthy motor strategies. Box plots show the median (50th percentile), interquartile range (25th--75th percentiles), whiskers extending to 1.5$\times$IQR, and outliers beyond this range.
    }
    \label{fig:hmm-err}
    \vspace{-2em}
\end{figure}
\begin{figure*}[htb]
    \centering
        \includegraphics[width=0.95\linewidth]{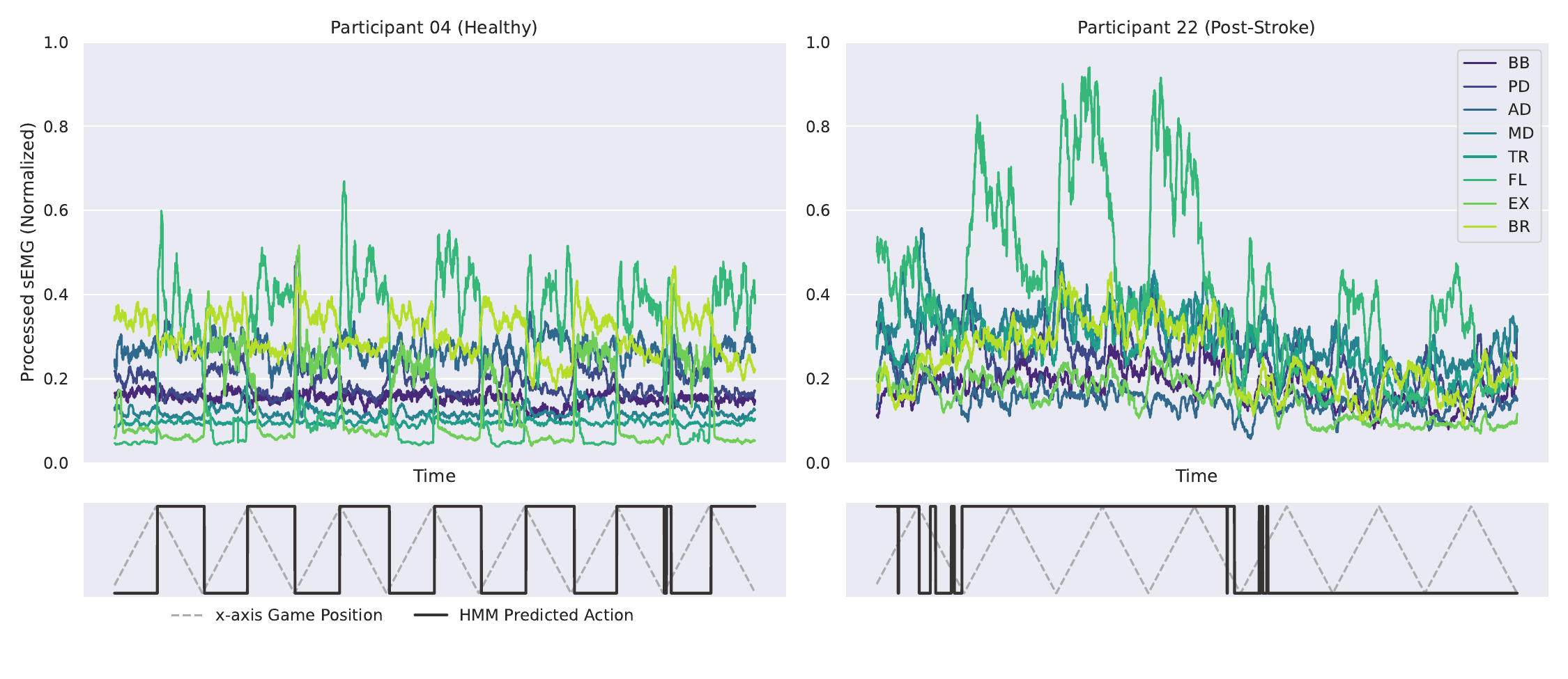}
        \vspace{-1em}
    \caption{Exemplar HMM-predicted action sequences (\emph{bottom}, \emph{solid black}), fit to 8-channel sEMG time series data (\emph{top}) as detailed in section~\ref{methods:HMM}, for the $x$-axis trajectory tracking task in pose Condition A, displayed alongside target $x$-axis game position (\emph{bottom}, \emph{dashed gray}) illustrating targeted cyclic horizontal direction changes. HMM-predicted action sequences consistently coincide with direction-specific subtasks for the healthy participant shown (04, \emph{left}) and for most healthy participants during most tasks, suggesting that directionality is the dominant factor determining underlying motor behavior. On the other hand, HMM-predicted action sequences substantially deviate from the target subtask sequence for the post-stroke participant shown (22, \emph{right}) and for both post-stroke participants during most tasks, indicating unmodeled complexity in neuromotor behaviors and/or additional time-varying factors impacting motor behavior. In this exemplar, HMM-identified actions correspond to temporal regions in which pathologically high wrist flexor activation was and was not present, indicating that HMM-based decompositions may be able to segment regions in which maladaptive compensatory behaviors are and are not expressed. Additional data on the distribution of HMM classification error across participants are available in appendix~V.
    }
    \label{fig:kl-div}
    \vspace{-2em}
\end{figure*}

\subsubsection{Hidden Markov model (HMM)--based decompositions enable discrimination of pathological neuromotor patterns in cyclic tasks}

As hypothesized in section~\ref{methods:HMM} and illustrated in Figure~\ref{fig:kl-div}, HMM-predicted action sequences consistently coincide with direction-specific subtasks in all 4 examined single-axis trajectory tracking tasks for the majority of healthy participants across both pose conditions. As also hypothesized and illustrated in Figure~\ref{fig:kl-div}, these correspondences break down for HMMs trained on post-stroke participant data, exposing qualitatively different underlying neuromotor behavior. As illustrated in Figure~\ref{fig:hmm-err}, our defined HMM subtask classification error metric quantitatively corroborates this difference: this error is consistently higher in magnitude and variance for post-stroke participants than for the vast majority of healthy individuals. The consistency of this finding --- which holds across a majority of examined tracking tasks and pose conditions --- strongly supports the potential utility of HMM-based decompositions for discriminating between healthy and pathological neuromotor behavior, and is especially noteworthy given that more standard synergy-based methods fail to capture any meaningful impairment-associated features in this data set.

Because HMMs are fit in a purely data-driven manner based on the statistical properties of the associated time series data, with no imposed physics- or physiology-grounded structure, interpreting precisely why these patterns are observed requires speculating on the underlying states the model is capturing. The consistent correspondence between direction-specific subtasks and HMM-identified actions when applied to healthy participant data suggests that directionality is the dominant factor determining underlying motor behavior for most healthy individuals performing most tasks. In contrast, the lack of 
this correspondence and the high variability of the defined error metric
in post-stroke participant data admits multiple interpretations. One possibility is that post-stroke participants are exhibiting more than 2 distinct motor behaviors, and HMMs assuming greater numbers of actions are needed to properly capture these additional modes. Another is that other time-varying factors influencing motor behavior are simply more dominant than directionality. It's even possible that, in some cases, the HMM is segmenting regions that do and do not exhibit compensatory behaviors, as illustrated in Figure~\ref{fig:kl-div}, providing a tool to track the expression of these behaviors over time. Interestingly, intermittently across tasks and pose conditions, some healthy participants --- such as participant 12 during the $x$-axis task, as shown in Figure~\ref{fig:hmm-err} --- also exhibit behavior for which directionality is not the dominant determinant of behavior modes, highlighting the challenge of discriminating pathological motor behavior from uncommon but healthy motor strategies even using metrics that effectively stratify the majority of individuals based on impairment level. Using such decompositions to design rehabilitation systems that impel users toward healthier behaviors will require deeper interrogation of these observed patterns to ground these results in physiological phenomena, but our successful use of these models to quantify pathology where more standard decompositions fail position HMMs --- and similar, more expressive probabilistic decompositions such as Gaussian mixture models --- as a promising tool to tease apart motor behavior during the complex, multidimensional movement tasks that are ultimately the target of neuromotor rehabilitation.

\vspace{-0.75em}
\subsection{Limitations}
The results above should be interpreted in the context of several key limitations.
First, all examined tasks were strictly isometric, with the robot fixed in space and forces mapped to avatar motion rather than physical limb displacement. While this enabled controlled interrogation of 6D force production and corresponding sEMG activations, the strictly isometric nature of the tasks limits the extent to which these findings generalize to dynamic, movement-based paradigms, which dominate most upper-limb rehabilitation systems and activities of daily living. Second, the post-stroke cohort was small ($n = 2$) and relatively homogeneous, consisting of individuals in the chronic phase with only mild spasticity and strength deficits. As such, the reported group differences --- particularly the negative findings in synergy analyses and the separability achieved with HMM-based methods --- must be interpreted as preliminary and may not extend to individuals with moderate-to-severe impairment or more heterogeneous lesion profiles. Third, although HMM-based decompositions demonstrated promising discriminatory capacity, the application of HMMs to upper-limb post-stroke neuromotor assessment remains sparse in the literature, limiting direct benchmarking against established clinical or biomechanical standards and necessitating further validation across larger datasets and task contexts. Fourth, the substantial inter-individual variability observed within the healthy cohort means even this larger sample size ($n=13$) does not admit definition of a singular ``normative'' motor profile, reducing the reliability of deviation-based pathology metrics and underscoring the challenge of distinguishing uncommon-but-healthy strategies from genuinely maladaptive ones. Finally, there remains limited foundational research on how specific task specifications --- such as force-to-visual mappings, dimensional constraints, and instruction framing --- systematically shape motor behavior in physical robot-mediated rehabilitation, making it challenging to benchmark our results. Consequently, some observed behavioral differences may reflect interface-induced adaptations rather than intrinsic pathology, reinforcing the need for controlled investigations of task design as an independent experimental factor.

\vspace{-0.75em}
\section{Conclusions and Future Work}
\label{sec:conc}

In aggregate, the findings above present a number of implications for the design of therapeutic robot interfaces, relevant to both our own planned OpenRobotRehab platform expansions and the wider rehabilitation community. First, our analyses illustrate that subtle aspects of system design --- many of which are not documented or reported on in the vast majority of studies --- can have outsize impact on user behavior, with major implications for the interpretation of neuromechanical data collected using any rehabilitation gaming interface, and for any associated assessment of motor pathology. Our specific findings highlighting the importance of explicitly constraining all motion dimensions you aim to control are actively informing our expansions of the OpenRobotRehab platform employed in this work (inspiring us, for example, to construct an augmented reality interface enabling the 3D display of target trajectories).

Second, the substantial heterogeneity we observe in healthy individuals' neuromotor behavior underscores the importance of incorporating healthy participants into study designs alongside neurologically impaired patient populations of interest to establish baselines and points of comparison, especially when data-driven models of pathology are pursued.
While it is by no means common practice, 
we aim to recruit large, diverse cohorts of individuals both with and without neuromotor pathology for all future studies.

Third, our success --- within the limited cohort of stroke participants and their relatively mild levels of impairment --- in classifying neuromotor pathology using HMM-based decomposition methods --- and corresponding failure to identify any such classifications using more standard synergy-based models --- illustrates both the exciting promise of probabilistic methods in quantifying neuromotor pathology and the extent to which ``healthy'' and ``unhealthy'' classifications of neuromotor behavior remain fundamentally mysterious. In the face of such foundational open questions --- and the numerous data-driven, simulation-based, and probabilistic approaches that aim to solve this problem --- we approach our own development of the OpenRobotRehab platform in a manner that is agnostic to the specific model of neuromotor pathology, seeking to enable ready comparison of existing approaches and adaptation to novel methodologies.

Lastly, comparison of our force- and HMM-based analyses underscores complementary strengths. Force-level metrics identify clear pathology-associated differences across heterogeneous 3D tasks using signals intrinsic to most rehabilitation robots, but primarily reflect performance and effort rather than underlying coordination. In contrast, within the constraints of our experimental cohort and tasks, the HMM-based decomposition reveals impairment-related temporal structure in sEMG that synergy methods do not detect, suggesting that probabilistic state-based models may more effectively capture latent neuromotor dynamics. Together, these results support the potential efficacy of rehabilitation platforms that combine mechanical performance measures with models of neuromotor organization while remaining agnostic to any single definition of ``healthy'' behavior.

Determining truly representative models of neuromotor pathology --- and robot rehabilitation systems that can make use of them to promote healthier neuromotor behaviors --- will require detailed, introspective analysis of observed user behaviors while accounting for the full system context we present in Figure~\ref{fig:sum}. We invite the wider rehabilitation community to contribute to such work alongside us --- empowered not only by the insights we present in this work, but by the OpenRobotRehab 1.0 data set we used to generate them, the first of many we aim to release as the OpenRobotRehab platform's capabilities continue to expand.

\vspace{-1em}
\section*{Acknowledgment}
The authors thank Carson J. Wynn, Evan Cole Falconer, and Jono Jenkens for their contributions to the development of the system that enabled data collection for this study.
\newpage
\bibliographystyle{IEEEtran}
\bibliography{ref.bib}

@inproceedings{anandextensible2025,
  title = {An {{Extensible Platform}} for {{Measurement}} and {{Modification}} of {{Muscle Engagement During Upper-Limb Robot-Facilitated Rehabilitation}}},
  booktitle = {2025 {{International Conference On Rehabilitation Robotics}} ({{ICORR}})},
  author = {Anand, Ajay and Berghoff, Chad A. and Wynn, Carson J. and Falconer, Evan Cole and Parra, Gabriel and Jenkens, Jono and Thomson, Caleb J. and Hamrick, W. Caden and George, Jacob A. and Hallock, Laura A.},
  year = {2025},
  pages = {1556--1563},
  issn = {1945-7901},
  doi = {10.1109/ICORR66766.2025.11062974},
  url = {https://ieeexplore.ieee.org/document/11062974},
  urldate = {2025-07-16},
  eventtitle = {2025 {{International Conference On Rehabilitation Robotics}} ({{ICORR}})}
}

@misc{who2024,
    author = "World Health Organization",
    title = "Rehabilitation Fact Sheet",
    year = "2024",
    howpublished = {\url{https://www.who.int/news-room/fact-sheets/detail/rehabilitation}},
    note = "[Online; accessed 2025-05-28]"
  }

@article{gassertRehabilitationRobotsTreatment2018,
  title = {Rehabilitation Robots for the Treatment of Sensorimotor Deficits: A Neurophysiological Perspective},
  shorttitle = {Rehabilitation Robots for the Treatment of Sensorimotor Deficits},
  author = {Gassert, Roger and Dietz, Volker},
  year = {2018},
  journal = {Journal of NeuroEngineering and Rehabilitation},
  shortjournal = {J NeuroEngineering Rehabil},
  volume = {15},
  number = {1},
  pages = {46},
  issn = {1743-0003},
  doi = {10.1186/s12984-018-0383-x},
  url = {https://doi.org/10.1186/s12984-018-0383-x},
  urldate = {2025-06-19},
  langid = {english}
}

@article{roseRoadForwardUpperextremity2021,
  title = {The Road Forward for Upper-Extremity Rehabilitation Robotics},
  author = {Rose, Chad G. and Deshpande, Ashish D. and Carducci, Jacob and Brown, Jeremy D.},
  year = {2021},
  journal = {Current Opinion in Biomedical Engineering},
  shortjournal = {Current Opinion in Biomedical Engineering},
  volume = {19},
  pages = {100291},
  issn = {2468-4511},
  doi = {10.1016/j.cobme.2021.100291},
  url = {https://www.sciencedirect.com/science/article/pii/S2468451121000325},
  urldate = {2025-06-19}
}

@article{salvalaggioPredictiveFactorsDose2024,
  title = {Predictive Factors and Dose–Response Effect of Rehabilitation for Upper Limb Induced Recovery after Stroke: Systematic Review with Proportional Meta-Analyses},
  shorttitle = {Predictive Factors and Dose–Response Effect of Rehabilitation for Upper Limb Induced Recovery after Stroke},
  author = {Salvalaggio, Silvia and Gianola, Silvia and Andò, Martina and Cacciante, Luisa and Castellini, Greta and Lando, Alex and Ossola, Gianluca and Pregnolato, Giorgia and Rutkowski, Sebastian and Vedovato, Anna and Zandonà, Chiara and Turolla, Andrea},
  year = {2024},
  journal = {Physiotherapy},
  shortjournal = {Physiotherapy},
  volume = {125},
  pages = {101417},
  issn = {0031-9406},
  doi = {10.1016/j.physio.2024.101417},
  url = {https://www.sciencedirect.com/science/article/pii/S0031940624004267},
  urldate = {2025-06-19}
}

@article{gauthierDoseResponseUpper2024,
  title = {Dose {{Response}} to {{Upper Extremity Stroke Rehabilitation Varies}} by {{Individual}}: {{Early Indicators}} of {{Treatment Response}}},
  shorttitle = {Dose {{Response}} to {{Upper Extremity Stroke Rehabilitation Varies}} by {{Individual}}},
  author = {Gauthier, Lynne V. and Ravi, Roshan and DeLuca, David and Zhou, Wenjin},
  year = {2024},
  journal = {Stroke},
  volume = {55},
  number = {3},
  pages = {696--704},
  publisher = {American Heart Association},
  doi = {10.1161/STROKEAHA.123.045039},
  url = {https://www.ahajournals.org/doi/full/10.1161/STROKEAHA.123.045039},
  urldate = {2025-06-19}
}

@article{moulaeiOverviewRoleRobots2023,
  title = {Overview of the Role of Robots in Upper Limb Disabilities Rehabilitation: A Scoping Review},
  shorttitle = {Overview of the Role of Robots in Upper Limb Disabilities Rehabilitation},
  author = {Moulaei, Khadijeh and Bahaadinbeigy, Kambiz and Haghdoostd, Ali Akbar and Nezhad, Mansour Shahabi and Sheikhtaheri, Abbas},
  year = {2023},
  journal = {Archives of Public Health},
  shortjournal = {Arch Public Health},
  volume = {81},
  number = {1},
  pages = {84},
  issn = {2049-3258},
  doi = {10.1186/s13690-023-01100-8},
  url = {https://doi.org/10.1186/s13690-023-01100-8},
  urldate = {2025-06-19},
  langid = {english}
}

@article{lohseMoreBetterUsing2014,
  title = {Is {{More Better}}? {{Using Metadata}} to {{Explore Dose}}–{{Response Relationships}} in {{Stroke Rehabilitation}}},
  shorttitle = {Is {{More Better}}?},
  author = {Lohse, Keith R. and Lang, Catherine E. and Boyd, Lara A.},
  year = {2014},
  journal = {Stroke},
  volume = {45},
  number = {7},
  pages = {2053--2058},
  publisher = {American Heart Association},
  doi = {10.1161/STROKEAHA.114.004695},
  url = {https://www.ahajournals.org/doi/full/10.1161/STROKEAHA.114.004695},
  urldate = {2025-06-19}
}

@article{langDoseResponseTaskspecific2016,
  title = {Dose Response of Task-Specific Upper Limb Training in People at Least 7 Months Poststroke: {{A}} Phase {{II}}, Single-Blind, Randomized, Controlled Trial},
  shorttitle = {Dose Response of Task-Specific Upper Limb Training in People at Least 6 Months Poststroke},
  author = {Lang, Catherine E. and Strube, Michael J. and Bland, Marghuretta D. and Waddell, Kimberly J. and Cherry-Allen, Kendra M. and Nudo, Randolph J. and Dromerick, Alexander W. and Birkenmeier, Rebecca L.},
  year = {2016},
  journal = {Annals of Neurology},
  volume = {80},
  number = {3},
  pages = {342--354},
  issn = {1531-8249},
  doi = {10.1002/ana.24734},
  url = {https://onlinelibrary.wiley.com/doi/abs/10.1002/ana.24734},
  urldate = {2025-07-16},
  langid = {english}
}

@article{pilaImpactDoseCombined2022a,
  title = {Impact of {{Dose}} of {{Combined Conventional}} and {{Robotic Therapy}} on {{Upper Limb Motor Impairments}} and {{Costs}} in {{Subacute Stroke Patients}}: {{A Retrospective Study}}},
  shorttitle = {Impact of {{Dose}} of {{Combined Conventional}} and {{Robotic Therapy}} on {{Upper Limb Motor Impairments}} and {{Costs}} in {{Subacute Stroke Patients}}},
  author = {Pila, OphÃ©lie and Koeppel, Typhaine and Grosmaire, Anne-GaÃ«lle and Duret, Christophe},
  year = {2022},
  journal = {Frontiers in Neurology},
  shortjournal = {Front Neurol},
  volume = {13},
  eprint = {35222240},
  eprinttype = {pubmed},
  pages = {770259},
  issn = {1664-2295},
  doi = {10.3389/fneur.2022.770259},
  url = {https://www.ncbi.nlm.nih.gov/pmc/articles/PMC8869251/},
  urldate = {2025-07-16},
  pmcid = {PMC8869251}
}

@article{sihMetabolicCostForce2003,
  title = {The {{Metabolic Cost}} of {{Force Generation}}},
  author = {Sih, Bryant L. and Stuhmiller, James H.},
  year = {2003},
  journal = {Medicine \& Science in Sports \& Exercise},
  volume = {35},
  number = {4},
  pages = {623--629},
  publisher = {Ovid Technologies (Wolters Kluwer Health)},
  issn = {0195-9131},
  doi = {10.1249/01.mss.0000058435.67376.49},
  url = {http://journals.lww.com/00005768-200304000-00013},
  urldate = {2025-07-16},
  langid = {english}
}

@article{kangForceControlChronic2015,
  title = {Force Control in Chronic Stroke},
  author = {Kang, Nyeonju and Cauraugh, James H.},
  year = {2015},
  journal = {Neuroscience \& Biobehavioral Reviews},
  shortjournal = {Neuroscience \& Biobehavioral Reviews},
  volume = {52},
  pages = {38--48},
  issn = {0149-7634},
  doi = {10.1016/j.neubiorev.2015.02.005},
  url = {https://www.sciencedirect.com/science/article/pii/S0149763415000500},
  urldate = {2025-07-16}
}

@article{seoAlterationsMotorModules2022a,
  title = {Alterations in Motor Modules and Their Contribution to Limitations in Force Control in the Upper Extremity after Stroke},
  author = {Seo, Gang and Lee, Sang Wook and Beer, Randall F. and Alamri, Amani and Wu, Yi-Ning and Raghavan, Preeti and Rymer, William Z. and Roh, Jinsook},
  year = {2022},
  journal = {Frontiers in Human Neuroscience},
  shortjournal = {Front. Hum. Neurosci.},
  volume = {16},
  publisher = {Frontiers},
  issn = {1662-5161},
  doi = {10.3389/fnhum.2022.937391},
  url = {https://www.frontiersin.org/journals/human-neuroscience/articles/10.3389/fnhum.2022.937391/full},
  urldate = {2025-07-16},
  langid = {english}
}

@article{rabinerTutorialHiddenMarkov1989,
  title={A tutorial on hidden Markov models and selected applications in speech recognition},
  author={Rabiner, Lawrence R},
  journal={Proceedings of the IEEE},
  volume={77},
  number={2},
  pages={257--286},
  year={2002},
  publisher={Ieee}
}

@article{Waskom2021,
    doi = {10.21105/joss.03021},
    url = {https://doi.org/10.21105/joss.03021},
    year = {2021},
    publisher = {The Open Journal},
    volume = {6},
    number = {60},
    pages = {3021},
    author = {Michael L. Waskom},
    title = {seaborn: statistical data visualization},
    journal = {Journal of Open Source Software}
 }

@article{latashBlissNotProblem2012,
  title = {The Bliss (Not the Problem) of Motor Abundance (Not Redundancy)},
  author = {Latash, Mark L.},
  year = {2012},
  journal = {Experimental Brain Research},
  shortjournal = {Exp Brain Res},
  volume = {217},
  number = {1},
  pages = {1--5},
  issn = {1432-1106},
  doi = {10.1007/s00221-012-3000-4},
  url = {https://doi.org/10.1007/s00221-012-3000-4},
  urldate = {2025-07-22},
  langid = {english}
}

@book{konradABCEMGPractical2005a,
  title = {The {{ABC}} of {{EMG}} -- A Practical Introduction to Kinesiological Electromyography},
  shorttitle = {The {{ABC}} of {{EMG}}},
  author = {Konrad, Peter},
  date = {2005-04},
  year = 2005,
  publisher = {Noraxon Inc. USA},
  langid = {english},
  pagetotal = {60}
}

@article{sobreperaAgeMotorFunction2025,
  title = {Age, {{Motor Function}}, and {{Cognitive Function Influence Preferences}} for {{Telerehabilitation Mediated}} by a {{Social Robot Augmented}} with {{Telepresence}}},
  author = {Sobrepera, Michael J. and Nguyen, Anh T. and Anand, Ajay and Prosser, Laura A. and Evans, Sally H. and Johnson, Michelle J.},
  year = {2025},
  journal = {IEEE transactions on neural systems and rehabilitation engineering: a publication of the IEEE Engineering in Medicine and Biology Society},
  shortjournal = {IEEE Trans Neural Syst Rehabil Eng},
  volume = {PP},
  eprint = {40699968},
  eprinttype = {pmid},
  issn = {1558-0210},
  doi = {10.1109/TNSRE.2025.3592020},
  langid = {english}
}

@article{alawiehFactorsAffectingPoststroke2018,
  title = {Factors Affecting Post-Stroke Motor Recovery: {{Implications}} on Neurotherapy after Brain Injury},
  shorttitle = {Factors Affecting Post-Stroke Motor Recovery},
  author = {Alawieh, Ali and Zhao, Jing and Feng, Wuwei},
  year = {2018},
  journal = {Behavioural brain research},
  shortjournal = {Behav Brain Res},
  volume = {340},
  eprint = {27531500},
  eprinttype = {pubmed},
  pages = {94--101},
  issn = {0166-4328},
  doi = {10.1016/j.bbr.2016.08.029},
  url = {https://www.ncbi.nlm.nih.gov/pmc/articles/PMC5305670/},
  urldate = {2025-07-31},
  pmcid = {PMC5305670}
}

@article{hatemRehabilitationMotorFunction2016,
  title={Rehabilitation of motor function after stroke: a multiple systematic review focused on techniques to stimulate upper extremity recovery},
  author={Hatem, Samar M and Saussez, Geoffroy and Della Faille, Margaux and Prist, Vincent and Zhang, Xue and Dispa, Delphine and Bleyenheuft, Yannick},
  journal={Frontiers in human neuroscience},
  volume={10},
  pages={442},
  year={2016},
  publisher={Frontiers Media SA}
}

@article{lodhaForceControlDegree2010,
  title = {Force Control and Degree of Motor Impairments in Chronic Stroke},
  author = {Lodha, Neha and Naik, Sagar K. and Coombes, Stephen A. and Cauraugh, James H.},
  year = {2010},
  journal = {Clinical Neurophysiology},
  shortjournal = {Clinical Neurophysiology},
  volume = {121},
  number = {11},
  pages = {1952--1961},
  issn = {1388-2457},
  doi = {10.1016/j.clinph.2010.04.005},
  url = {https://www.sciencedirect.com/science/article/pii/S1388245710003561},
  urldate = {2025-08-01}}

@article{ortega-auriolRoleMuscleSynergies2025,
  title = {The Role of Muscle Synergies and Task Constraints on Upper Limb Motor Impairment after Stroke},
  author = {Ortega-Auriol, Pablo and Byblow, Winston D. and Ren, April Xiaoge and Besier, Thor and McMorland, Angus J. C.},
  year = {2025},
  journal = {Experimental Brain Research},
  shortjournal = {Exp Brain Res},
  volume = {243},
  number = {1},
  pages = {40},
  issn = {1432-1106},
  doi = {10.1007/s00221-024-06953-1},
  url = {https://doi.org/10.1007/s00221-024-06953-1},
  urldate = {2025-08-01},
  langid = {english}
}

@article{rohAlterationsUpperLimb2013,
  title = {Alterations in Upper Limb Muscle Synergy Structure in Chronic Stroke Survivors},
  author = {Roh, Jinsook and Rymer, William Z. and Perreault, Eric J. and Yoo, Seng Bum and Beer, Randall F.},
  year = {2013},
  journal = {Journal of Neurophysiology},
  volume = {109},
  number = {3},
  pages = {768--781},
  publisher = {American Physiological Society},
  issn = {0022-3077},
  doi = {10.1152/jn.00670.2012},
  url = {https://journals.physiology.org/doi/full/10.1152/jn.00670.2012},
  urldate = {2025-03-18}
}

@article{rohEvidenceAlteredUpper2015,
  title = {Evidence for Altered Upper Extremity Muscle Synergies in Chronic Stroke Survivors with Mild and Moderate Impairment},
  author = {Roh, Jinsook and Rymer, William Z. and Beer, Randall F.},
  year = {2015},
  journal = {Frontiers in Human Neuroscience},
  shortjournal = {Front Hum Neurosci},
  volume = {9},
  eprint = {25717296},
  eprinttype = {pubmed},
  pages = {6},
  issn = {1662-5161},
  doi = {10.3389/fnhum.2015.00006},
  langid = {english},
  pmcid = {PMC4324145}
}

@article{clarkMergingHealthyMotor2010,
  title = {Merging of Healthy Motor Modules Predicts Reduced Locomotor Performance and Muscle Coordination Complexity Post-Stroke},
  author = {Clark, David J. and Ting, Lena H. and Zajac, Felix E. and Neptune, Richard R. and Kautz, Steven A.},
  year = {2010},
  journal = {Journal of Neurophysiology},
  shortjournal = {J Neurophysiol},
  volume = {103},
  number = {2},
  eprint = {20007501},
  eprinttype = {pubmed},
  pages = {844--857},
  issn = {1522-1598},
  doi = {10.1152/jn.00825.2009},
  langid = {english},
  pmcid = {PMC2822696}
}

@article{davellaControlFastReachingMovements2006,
  title={Control of fast-reaching movements by muscle synergy combinations},
  author={d'Avella, Andrea and Portone, Alessandro and Fernandez, Laure and Lacquaniti, Francesco},
  journal={Journal of Neuroscience},
  volume={26},
  number={30},
  pages={7791--7810},
  year={2006},
  publisher={Society for Neuroscience}
}

@article{scanoMuscleSynergiesBasedCharacterization2017,
  title = {Muscle {{Synergies-Based Characterization}} and {{Clustering}} of {{Poststroke Patients}} in {{Reaching Movements}}},
  author = {Scano, Alessandro and Chiavenna, Andrea and Malosio, Matteo and Molinari Tosatti, Lorenzo and Molteni, Franco},
  year = {2017},
  journal = {Frontiers in Bioengineering and Biotechnology},
  shortjournal = {Front. Bioeng. Biotechnol.},
  volume = {5},
  publisher = {Frontiers},
  issn = {2296-4185},
  doi = {10.3389/fbioe.2017.00062},
  url = {https://www.frontiersin.org/journals/bioengineering-and-biotechnology/articles/10.3389/fbioe.2017.00062/full},
  urldate = {2025-08-01},
  langid = {english}
}

@article{derugyAreMuscleSynergies2013,
  title={Are muscle synergies useful for neural control?},
  author={de Rugy, Aymar and Loeb, Gerald E and Carroll, Timothy J},
  journal={Frontiers in computational neuroscience},
  volume={7},
  pages={19},
  year={2013},
  publisher={Frontiers Media SA}
}

@article{treschCaseMuscleSynergies2009,
  title = {The Case for and against Muscle Synergies},
  author = {Tresch, Matthew C and Jarc, Anthony},
  year = {2009},
  journal = {Current Opinion in Neurobiology},
  shortjournal = {Current Opinion in Neurobiology},
  series = {Motor Systems â€¢ {{Neurology}} of Behaviour},
  volume = {19},
  number = {6},
  pages = {601--607},
  issn = {0959-4388},
  doi = {10.1016/j.conb.2009.09.002},
  url = {https://www.sciencedirect.com/science/article/pii/S095943880900124X},
  urldate = {2025-08-01}
}

@article{reinboltSimulationHumanMovement2011,
  title = {Simulation of Human Movement: Applications Using {{OpenSim}}},
  shorttitle = {Simulation of Human Movement},
  author = {Reinbolt, Jeffrey A. and Seth, Ajay and Delp, Scott L.},
  year = {2011},
  journal = {Procedia IUTAM},
  shortjournal = {Procedia IUTAM},
  series = {{{IUTAM Symposium}} on {{Human Body Dynamics}}},
  volume = {2},
  pages = {186--198},
  issn = {2210-9838},
  doi = {10.1016/j.piutam.2011.04.019},
  url = {https://www.sciencedirect.com/science/article/pii/S2210983811000204},
  urldate = {2025-08-01}
}

@article{sethOpenSimSimulatingMusculoskeletal2018,
  title = {{{OpenSim}}: {{Simulating}} Musculoskeletal Dynamics and Neuromuscular Control to Study Human and Animal Movement},
  shorttitle = {{{OpenSim}}},
  author = {Seth, Ajay and Hicks, Jennifer L. and Uchida, Thomas K. and Habib, Ayman and Dembia, Christopher L. and Dunne, James J. and Ong, Carmichael F. and DeMers, Matthew S. and Rajagopal, Apoorva and Millard, Matthew and Hamner, Samuel R. and Arnold, Edith M. and Yong, Jennifer R. and Lakshmikanth, Shrinidhi K. and Sherman, Michael A. and Ku, Joy P. and Delp, Scott L.},
  year = {2018},
  journal = {PLOS Computational Biology},
  shortjournal = {PLOS Computational Biology},
  volume = {14},
  number = {7},
  pages = {e1006223},
  publisher = {Public Library of Science},
  issn = {1553-7358},
  doi = {10.1371/journal.pcbi.1006223},
  url = {https://journals.plos.org/ploscompbiol/article?id=10.1371/journal.pcbi.1006223},
  urldate = {2025-08-01},
  langid = {english}
}

@article{harringtonMusculoskeletalModelingMovement2024,
  title = {Musculoskeletal Modeling and Movement Simulation for Structural Hip Disorder Research: {{A}} Scoping Review of Methods, Validation, and Applications},
  shorttitle = {Musculoskeletal Modeling and Movement Simulation for Structural Hip Disorder Research},
  author = {Harrington, Margaret S. and Leo, Stefania D. F. Di and Hlady, Courtney A. and Burkhart, Timothy A.},
  year = {2024},
  journal = {Heliyon},
  shortjournal = {Heliyon},
  volume = {10},
  number = {15},
  eprint = {39157349},
  eprinttype = {pubmed},
  publisher = {Elsevier},
  issn = {2405-8440},
  doi = {10.1016/j.heliyon.2024.e35007},
  url = {https://www.cell.com/heliyon/abstract/S2405-8440(24)11038-9},
  urldate = {2025-08-01},
  langid = {english}
}

@inproceedings{anwer2022rehabilitation,
  title={Rehabilitation of Upper Limb Motor Impairment in Stroke: A Narrative Review on the Prevalence, Risk Factors, and Economic Statistics of Stroke and State of the Art Therapies},
  author={Anwer, Saba and Waris, Asim and Gilani, Syed Omer and Iqbal, Javaid and Shaikh, Nusratnaaz and Pujari, Amit N and Niazi, Imran Khan},
  booktitle={Healthcare},
  volume={10},
  number={2},
  pages={190},
  year={2022},
  organization={MDPI}
}

@article{blank2014current,
  title={Current trends in robot-assisted upper-limb stroke rehabilitation: promoting patient engagement in therapy},
  author={Blank, Amy A and French, James A and Pehlivan, Ali Utku and O’Malley, Marcia K},
  journal={Current physical medicine and rehabilitation reports},
  volume={2},
  number={3},
  pages={184--195},
  year={2014},
  publisher={Springer}
}

@article{LIN2015946,
title = {Occupational Therapy Workforce in the {U}nited {S}tates: Forecasting Nationwide Shortages},
journal = {PM\&R},
volume = {7},
number = {9},
pages = {946-954},
year = {2015},
issn = {1934-1482},
doi = {https://doi.org/10.1016/j.pmrj.2015.02.012},
url = {https://www.sciencedirect.com/science/article/pii/S1934148215000866},
author = {Vernon Lin and Xiaoming Zhang and Pamela Dixon}
}

@INPROCEEDINGS{pezzera2020,

  author={Pezzera, Manuel and Borghese, N. Alberto},

  booktitle={2020 IEEE 8th International Conference on Serious Games and Applications for Health (SeGAH)}, 

  title={Dynamic difficulty adjustment in exer-games for rehabilitation: a mixed approach}, 

  year={2020},

  volume={},

  number={},

  pages={1-7},

  doi={10.1109/SeGAH49190.2020.9201871}}

@article{funato2022muscle,
  title={Muscle synergy analysis yields an efficient and physiologically relevant method of assessing stroke},
  author={Funato, Tetsuro and Hattori, Noriaki and Yozu, Arito and An, Qi and Oya, Tomomichi and Shirafuji, Shouhei and Jino, Akihiro and Miura, Kyoichi and Martino, Giovanni and Berger, Denise and others},
  journal={Brain Communications},
  volume={4},
  number={4},
  pages={fcac200},
  year={2022},
  publisher={Oxford University Press}
}

@article{berger2024myoelectric,
  title={Myoelectric control and virtual reality to enhance motor rehabilitation after stroke},
  author={Berger, Denise Jennifer and d’Avella, Andrea},
  journal={Frontiers in Bioengineering and Biotechnology},
  volume={12},
  pages={1376000},
  year={2024},
  publisher={Frontiers Media SA}
}

@article{delp2007opensim,
  title={Open{S}im: open-source software to create and analyze dynamic simulations of movement},
  author={Delp, Scott L and Anderson, Frank C and Arnold, Allison S and Loan, Peter and Habib, Ayman and John, Chand T and Guendelman, Eran and Thelen, Darryl G},
  journal={IEEE Transactions on Biomedical Engineering},
  volume={54},
  number={11},
  pages={1940--1950},
  year={2007},
  publisher={IEEE}
}

@article{caggiano2022myosuite,
  title={Myo{S}uite--{A} contact-rich simulation suite for musculoskeletal motor control},
  author={Caggiano, Vittorio and Wang, Huawei and Durandau, Guillaume and Sartori, Massimo and Kumar, Vikash},
  journal={arXiv preprint arXiv:2205.13600},
  year={2022}
}

@article{park2023relevance,
  title={Relevance of Upper Limb Muscle Synergies to Dynamic Force Generation: Perspectives on Rehabilitation of Impaired Intermuscular Coordination in Stroke},
  author={Park, Jeong-Ho and Shin, Joon-Ho and Lee, Hangil and Roh, Jinsook and Park, Hyung-Soon},
  journal={IEEE Transactions on Neural Systems and Rehabilitation Engineering},
  volume={31},
  pages={4851--4861},
  year={2023},
  publisher={IEEE}
}

@book{valero2016fundamentals,
  title={Fundamentals of Neuromechanics},
  author={Valero-Cuevas, Francisco J},
  volume={8},
  year={2016},
  publisher={Springer}
}

@article{park2024effects,
  title={Effects of Robot-Assisted Therapy for Upper Limb Rehabilitation After Stroke: An Umbrella Review of Systematic Reviews},
  author={Park, Jong Mi and Park, Hee Jae and Yoon, Seo Yeon and Kim, Yong Wook and Shin, Jae Il and Lee, Sang Chul},
  journal={Stroke},
  year={2024},
  publisher={Lippincott Williams \& Wilkins Hagerstown, MD}
}

@article{wei2024systematic,
  title={A systematic review and meta-analysis of clinical efficacy of early and late rehabilitation interventions for ischemic stroke},
  author={Wei, Xufang and Sun, Shengtong and Zhang, Manyu and Zhao, Zhenqiang},
  journal={BMC neurology},
  volume={24},
  number={1},
  pages={91},
  year={2024},
  publisher={Springer}
}

@article{yagi2017impact,
  title={Impact of rehabilitation on outcomes in patients with ischemic stroke: a nationwide retrospective cohort study in Japan},
  author={Yagi, Maiko and Yasunaga, Hideo and Matsui, Hiroki and Morita, Kojiro and Fushimi, Kiyohide and Fujimoto, Masashi and Koyama, Teruyuki and Fujitani, Junko},
  journal={Stroke},
  volume={48},
  number={3},
  pages={740--746},
  year={2017},
  publisher={Lippincott Williams \& Wilkins Hagerstown, MD}
}

@article{ellis2009impairment,
  title={Impairment-based 3-D robotic intervention improves upper extremity work area in chronic stroke: targeting abnormal joint torque coupling with progressive shoulder abduction loading},
  author={Ellis, Michael D and Sukal-Moulton, Theresa M and Dewald, Julius PA},
  journal={IEEE Transactions on Robotics},
  volume={25},
  number={3},
  pages={549--555},
  year={2009},
  publisher={IEEE}
}

@article{ellis2009progressive,
  title={Progressive shoulder abduction loading is a crucial element of arm rehabilitation in chronic stroke},
  author={Ellis, Michael D and Sukal-Moulton, Theresa and Dewald, Julius PA},
  journal={Neurorehabilitation and neural repair},
  volume={23},
  number={8},
  pages={862--869},
  year={2009},
  publisher={SAGE Publications Sage CA: Los Angeles, CA}
}

@article{ellis2018progressive,
  title={Progressive abduction loading therapy with horizontal-plane viscous resistance targeting weakness and flexion synergy to treat upper limb function in chronic hemiparetic stroke: a randomized clinical trial},
  author={Ellis, Michael D and Carmona, Carolina and Drogos, Justin and Dewald, Julius PA},
  journal={Frontiers in neurology},
  volume={9},
  pages={71},
  year={2018},
  publisher={Frontiers Media SA}
}

@article{bizzi2013neural,
  title={The neural origin of muscle synergies},
  author={Bizzi, Emilio and Cheung, Vincent CK},
  journal={Frontiers in computational neuroscience},
  volume={7},
  pages={51},
  year={2013},
  publisher={Frontiers Media SA}
}

@article{latash2024useful,
  title={Useful and useless misnomers in motor control},
  author={Latash, Mark L},
  journal={Motor control},
  volume={29},
  number={1},
  pages={69--98},
  year={2024},
  publisher={Human Kinetics}
}

@article{bohannonInterraterReliabilityModified1987,
  title = {Interrater Reliability of a Modified {A}shworth Scale of Muscle Spasticity},
  author = {Bohannon, Richard W. and Smith, Melissa B.},
  year = {1987},
  journal = {Physical Therapy},
  shortjournal = {Physical Therapy},
  volume = {67},
  number = {2},
  pages = {206--207},
  issn = {0031-9023},
  doi = {10.1093/ptj/67.2.206},
  ourl = {https://doi.org/10.1093/ptj/67.2.206},
  urldate = {2024-11-22}
}

@article{ManualMuscleTesting,
  title = {Manual Muscle Testing},
  year = {1990},
  author = {Mendell, J.R. and Florence, J. },
  journal = {Muscle \& Nerve},
  volume = {13},
  pages = {S16-S20},
  ourl = {https://onlinelibrary.wiley.com/doi/abs/10.1002/mus.880131307},
  urldate = {2024-11-22}
}

@article{alessandroMuscleSynergiesNeuroscience2013,
  title={Muscle synergies in neuroscience and robotics: From input-space to task-space perspectives},
  author={Alessandro, Cristiano and Delis, Ioannis and Nori, Francesco and Panzeri, Stefano and Berret, Bastien},
  journal={Frontiers in Computational Neuroscience},
  volume={7},
  pages={43},
  year={2013},
  publisher={Frontiers Media SA}
}

@article{dipietroChangingMotorSynergies2007,
  title={Changing motor synergies in chronic stroke},
  author={Dipietro, Laura and Krebs, Hermano I and Fasoli, Susan E and Volpe, Bruce T and Stein, Joel and Bever, C and Hogan, Neville},
  journal={Journal of Neurophysiology},
  volume={98},
  number={2},
  pages={757--768},
  year={2007},
  publisher={American Physiological Society}
}

@article{facciorussoMuscleSynergiesUpper2024,
  title={Muscle synergies in upper limb stroke rehabilitation: A
 scoping review},
  author={Facciorusso, Salvatore and Guanziroli, Eleonora and Brambilla, Cristina and Spina, Stefania and Giraud, Manuela and Tosatti, Lorenzo Molinari and Santamato, Andrea and Molteni, Franco and Scano, Alessandro},
  journal={European Journal of Physical and Rehabilitation Medicine},
  volume={60},
  number={5},
  pages={767},
  year={2024}
}

@ARTICLE{1450960,
  author={Forney, G.D.},
  journal={Proceedings of the IEEE}, 
  title={The viterbi algorithm}, 
  year={1973},
  volume={61},
  number={3},
  pages={268-278},
  doi={10.1109/PROC.1973.9030}
}

@ARTICLE{4520143,
  author={Chiang, Joyce and Wang, Z. Jane and McKeown, Martin J.},
  journal={IEEE Transactions on Signal Processing}, 
  title={A Hidden Markov, Multivariate Autoregressive (HMM-mAR) Network Framework for Analysis of Surface EMG (sEMG) Data}, 
  year={2008},
  volume={56},
  number={8},
  pages={4069-4081},
  doi={10.1109/TSP.2008.925246}
}

@ARTICLE{7839943,
  author={Razin, Yosef S. and Pluckter, Kevin and Ueda, Jun and Feigh, Karen},
  journal={IEEE Robotics and Automation Letters}, 
  title={Predicting Task Intent From Surface Electromyography Using Layered Hidden Markov Models}, 
  year={2017},
  volume={2},
  number={2},
  pages={1180-1185},
  doi={10.1109/LRA.2017.2662741}
}

@article{WEN2021102592,
title = {Human hand movement recognition using infinite hidden Markov model based sEMG classification},
journal = {Biomedical Signal Processing and Control},
volume = {68},
pages = {102592},
year = {2021},
issn = {1746-8094},
doi = {https://doi.org/10.1016/j.bspc.2021.102592},
url = {https://www.sciencedirect.com/science/article/pii/S1746809421001890},
author = {Ruoshi Wen and Qiang Wang and Zhibin Li}
}

\end{document}


\title{Appendices: Characterizing Healthy \& Post-Stroke Neuromotor Behavior During 6D Upper-Limb Isometric Gaming: Implications for Design of End-Effector Rehabilitation Robot Interfaces} 

\author{Ajay Anand, \IEEEmembership{Member, IEEE}, Gabriel Parra, Chad A. Berghoff, and Laura A. Hallock,
\IEEEmembership{Member, IEEE}}

\maketitle

\section{Ideal Force Calculation}
\label{supp:ideal}
To estimate the ideal force profiles corresponding to perfect task execution across the 7 force production tasks, we utilize the constant in-game scaling factor --- fixed throughout the entire experiment --- that maps end-effector forces to the avatar’s velocity in the virtual environment. For each participant, measured avatar velocities are regressed against this known scaling factor to estimate constant force offsets along the $x$, $y$, and $z$ axes. These offsets, which remain consistent across all seven tasks within a given experimental condition, account for baseline measurement biases introduced by the handle–sensor interface and configuration-dependent changes in the robot system. After determining these offsets, the recorded position of the dynamic target is used to compute the theoretical force that would have been required to guide the avatar precisely to overlap with the target as it progresses through each trajectory task, thereby representing perfect task completion. These idealized force values are subsequently used as the reference for all root mean squared error (RMSE) calculations.

\section{$K$-means Clustering Procedures for Decomposed Synergies}
\label{supp:synergy}

We explored the following techniques to categorize decomposed synergies using $k$-means clustering into groups of participants with similar synergies:
\begin{enumerate}
    \item $K$-means was initialized with cluster counts 1--8, and synergies labeled with their respective participants were then assigned a cluster. We also initialized $k$-means with cluster counts greater than 8 and observed similiar results.
    \item $K$-means was initialized with cluster counts 1--8. Participants' top-$n$ synergies were then assigned clusters, and each participant was then clustered using the labels assigned to their respective synergies. 
    \item The dimensionality of synergies makes it difficult to accurately classify participants, as multiple synergies decomposed from each participant may belong to a different cluster. To simplify the classification of participants, we concatenated each participant's synergies (3--5 vectors in $R^8$) into a single vector (in $R^{15}$--$R^{40}$). $K$-means was then used to cluster the concatenated synergy vector, constraining participants to belong exclusively to a single group.
\end{enumerate}

All of these clustering techniques failed to categorize decomposed participant synergies into meaningful groups.

\section{Hidden Markov Model (HMM) Subtask Classification Error}
\label{supp:hmm}

We define the subtask classification error of an HMM fitted to the time series sEMG data collected during performance of a given trajectory tracking task as follows.
Scoring relies on three key trajectories, each with elements in the set $\{0, 1\}$. The first trajectory $V = \{x_t\}^T_{t=1}$ consists of actions generated by an HMM. The second trajectory $A = \{a_t\}^T_{t=1}$ consists of the subtasks prescribed at time $t$. The third trajectory $A^C = f(A)$ where $f: 0 \mapsto 1$ and $f: 1 \mapsto 0$ is constructed to account for the fact that HMMs assign random labels on initialization, allowing us to ``swap'' state classification.
The subtask classification error is defined as:
$$
    S(A, A^C, V) = \min{ \{ \text{hamming}(A, V), \text{hamming}(A^C, V) \}}
$$
where Hamming loss is used to give a percentage of subtasks that are misidentified. Thus, the model's scores improve as they approach $0.0$ and worsen as they approach $0.5$ (random chance).

\newpage

\section {Condition B Force Aggregates}
\label {supp:ConB}

\begin{figure}[htb]
\centering

\subfloat[]{%
\includegraphics[width=0.22\textwidth]{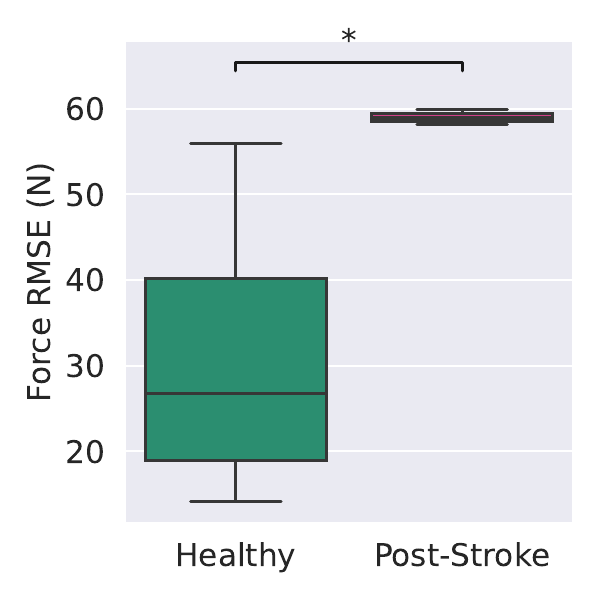}
\label{fig:forcesRMSE-B}
}
\hfill
\subfloat[]{%
\includegraphics[width=0.22\textwidth]{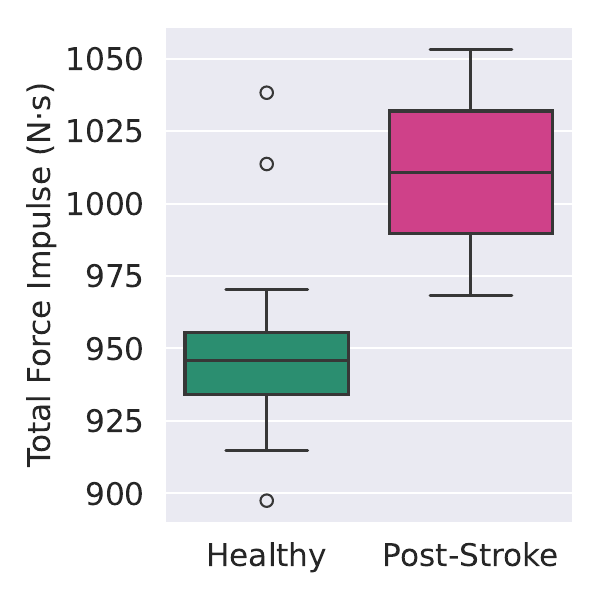}
\label{fig:forcesImpulses-B}
}
\hfill
\subfloat[]{%
\includegraphics[width=0.22\textwidth]{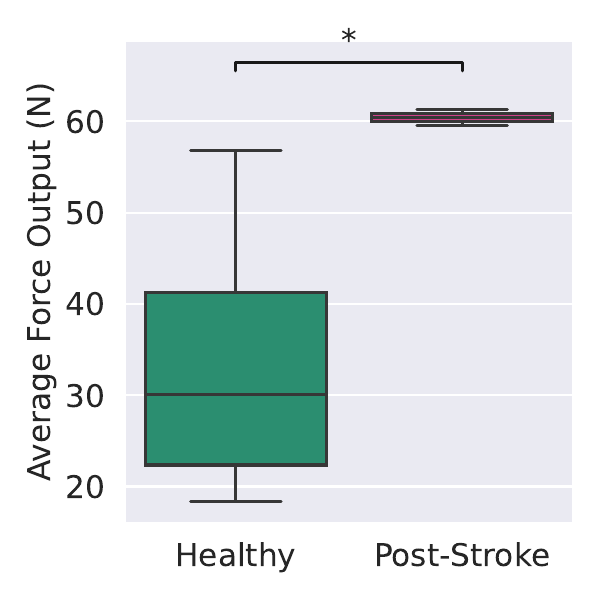}
\label{fig:forcesRMS-B}
}
\hfill
\subfloat[]{%
\includegraphics[width=0.22\textwidth]{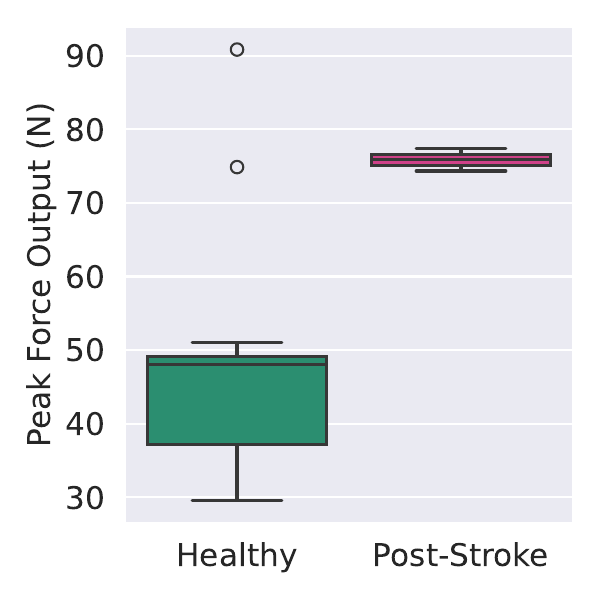}
\label{fig:forcesPeak-B}
}

\caption{``Productive'' force RMSE (a), total force impulse (b), average force output (c), and peak force output (d), as calculated for each of the 7 force trajectory tracking tasks at pose Condition B, aggregated across all healthy (02--20, \emph{turquoise}) and post-stroke (21--22, \emph{pink}) participants. Post-stroke participants exhibit significantly higher force RMSE, consistent with existing literature, as well as significantly higher average force output. Total force impulse and peak force output appeared higher for impaired participants, but results were not statistically significant. These findings were consistent across both pose conditions. Box plots show the median (50th percentile), interquartile range (25th--75th percentiles), whiskers extending to 1.5$\times$IQR, and outliers beyond this range.}
\label{fig:forcesAgg-B}
\end{figure}

\section{Hidden Markov Model (HMM) Subtask Classification Loss Distributions}
\label{supp:hmm-loss}

\begin{figure}[htb]
\centering
\subfloat{%
\includegraphics[width=0.22\textwidth]{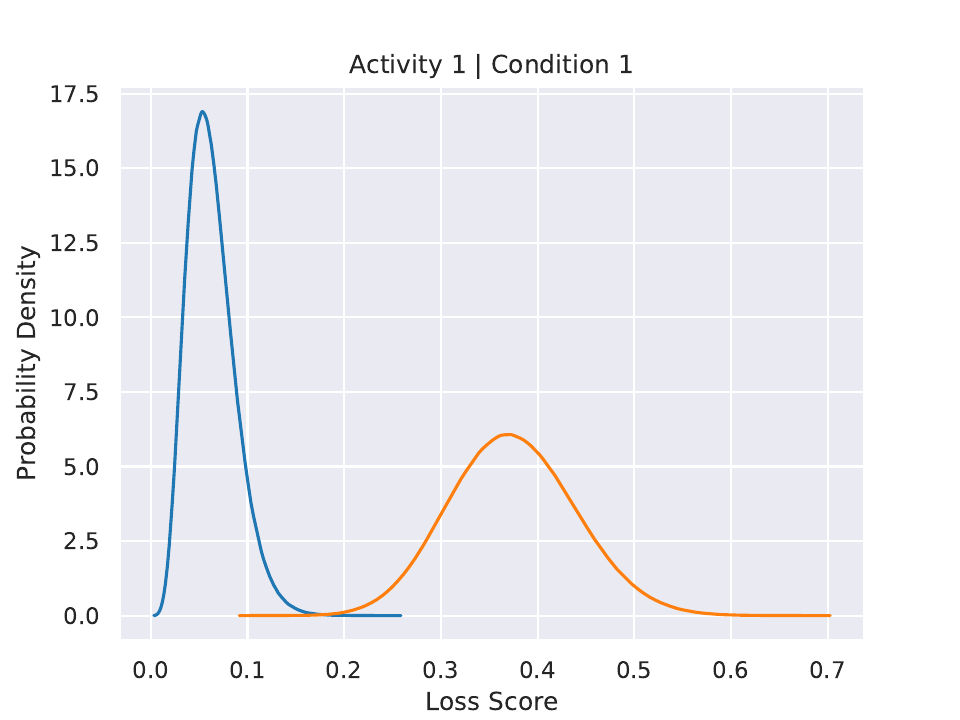}
}
\subfloat{%
\includegraphics[width=0.22\textwidth]{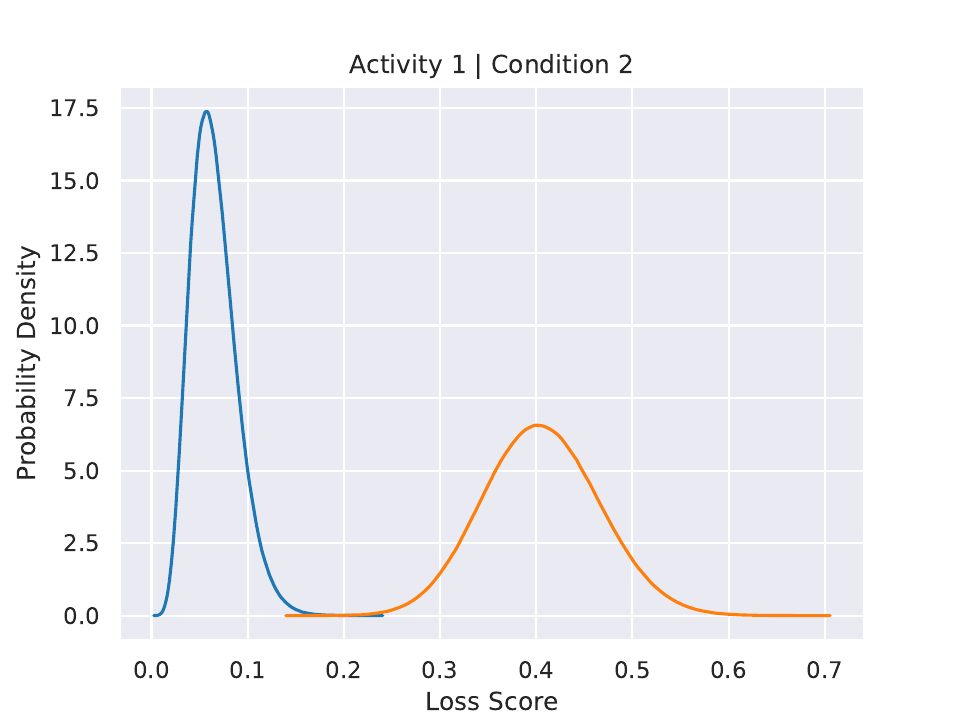}
}
\subfloat{%
\includegraphics[width=0.22\textwidth]{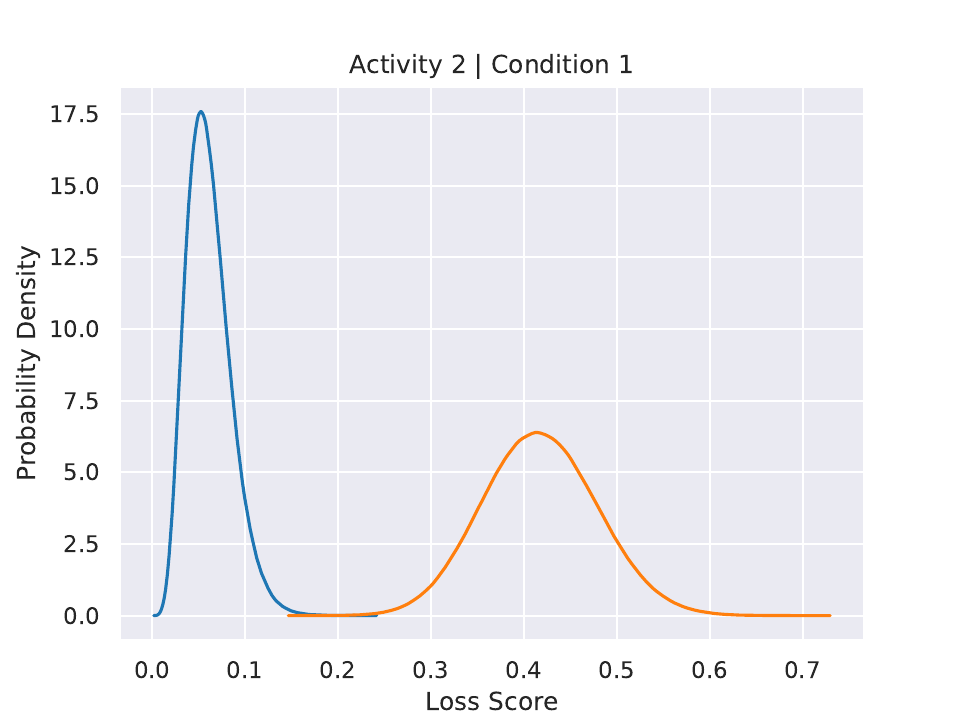}
}
\subfloat{%
\includegraphics[width=0.22\textwidth]{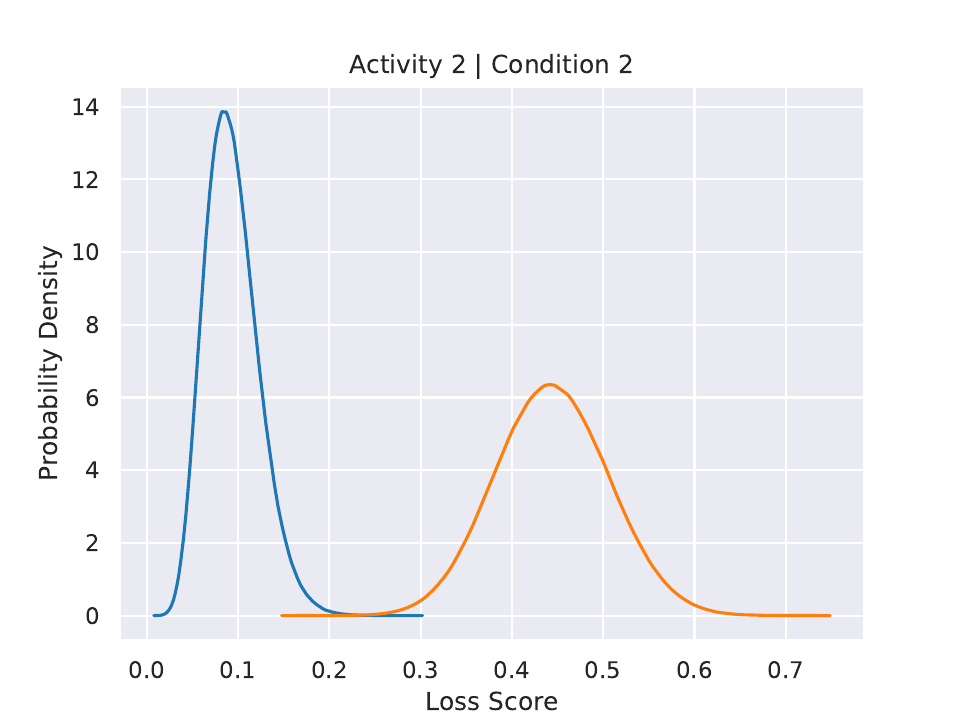}
}

\subfloat{%
\includegraphics[width=0.22\textwidth]{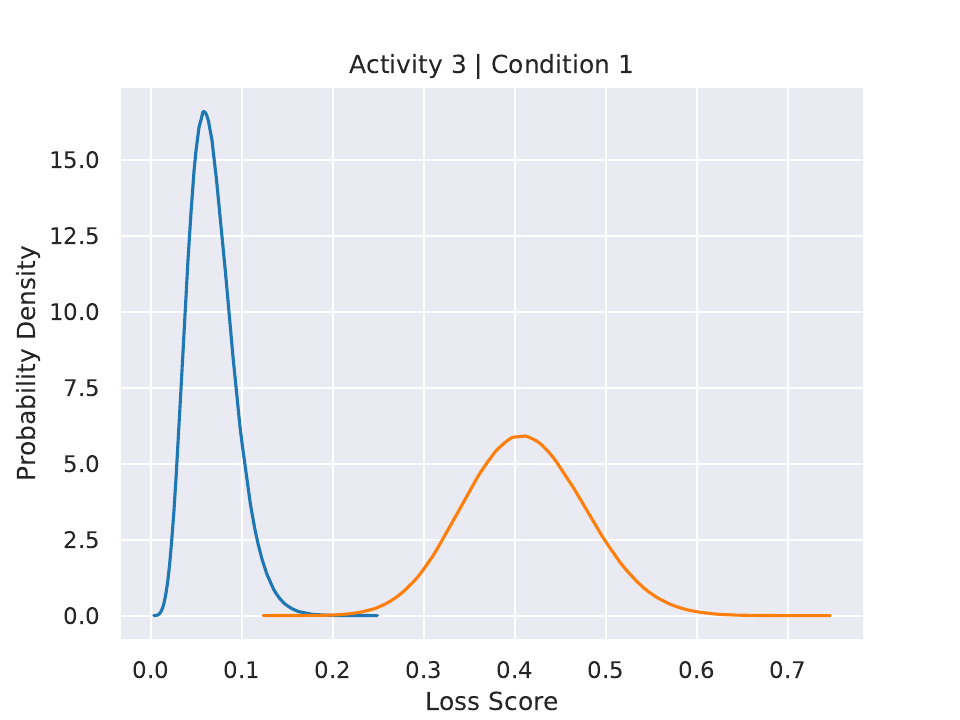}
}
\subfloat{%
\includegraphics[width=0.22\textwidth]{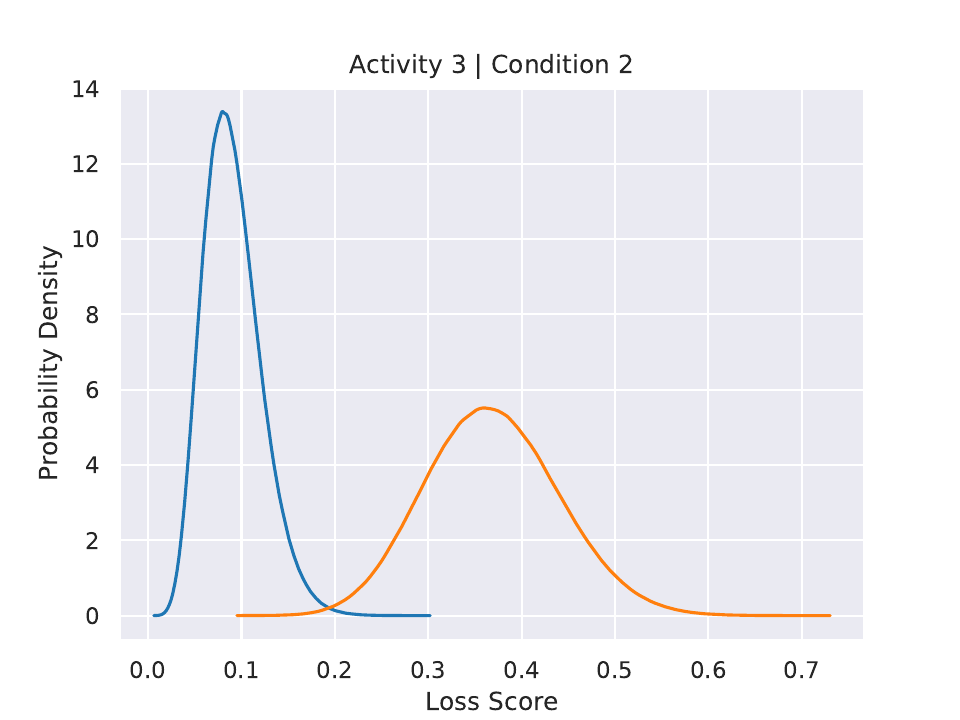}
}
\subfloat{%
\includegraphics[width=0.22\textwidth]{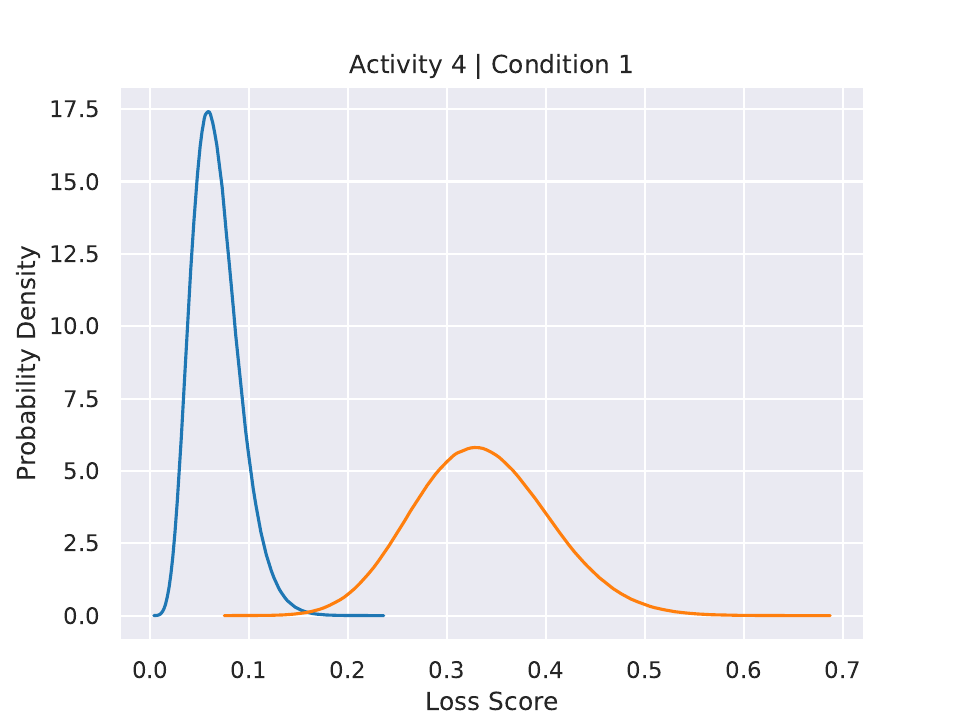}
}
\subfloat{%
\includegraphics[width=0.22\textwidth]{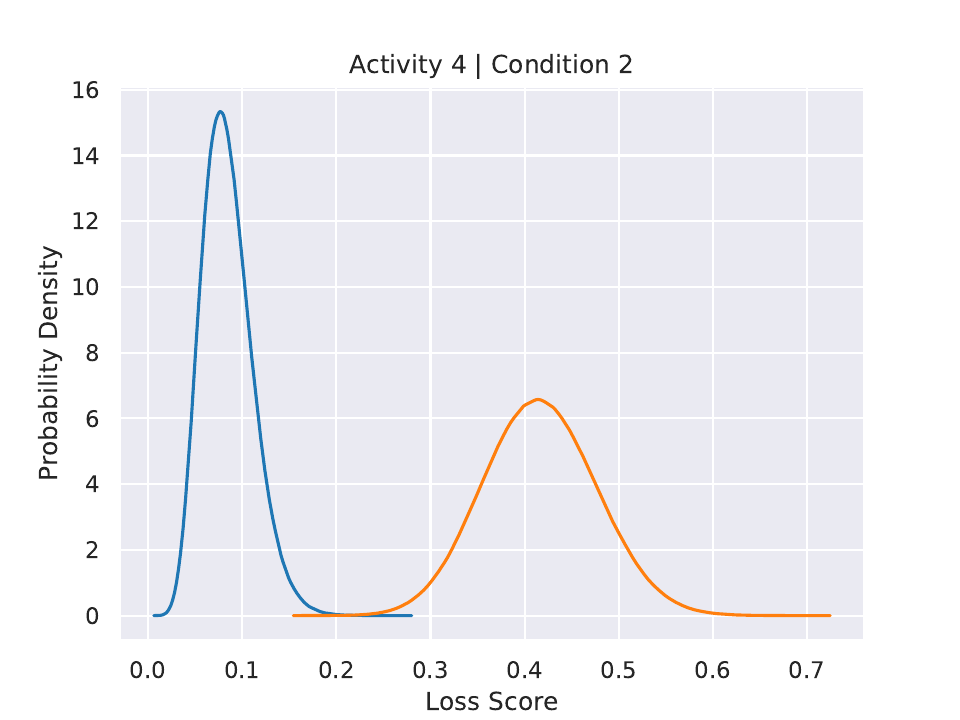}
}
\caption{Participants falling into non-normative (\emph{orange}) --- as contrasted with normative (\emph{blue}) --- loss score distribution. Post-stroke participants (21, 22) or participants who were observed to have atypical behavior (8) are the only participants who fell into the non-normative distribution on half or more of the tasks.
}
\end{figure}

\begin{table}[h]
\centering
\caption{Participants in Non-Normative Loss Distribution}
\begin{tabular}{|l|l|l|}
\hline
\textbf{Task} & \textbf{Condition} & \textbf{Participants in Non-Normative Loss Distribution} \\
\hline
$x$-axis & Condition 1 & 8, 12, 13, 15, 21, 22 \\
\hline
$x$-axis & Condition 2 & 12, 21, 22 \\
\hline
$y$-axis & Condition 1 & 3, 5, 21, 22 \\
\hline
$y$-axis & Condition 2 & 3, 15 \\
\hline
$z$-axis & Condition 1 & 20, 22 \\
\hline
$z$-axis & Condition 2 & 7, 8, 13, 22 \\
\hline
torque & Condition 1 & 7, 8, 12, 13 21 \\
\hline
torque & Condition 2 & 8, 21, 22 \\
\hline
\end{tabular}
\vspace{1em}
\label{tab:my_table}
\end{table}